\documentclass[sigconf, nonacm, screen=true]{acmart}

\newcommand\vldbdoi{XX.XX/XXX.XX}

\newcommand\vldbavailabilityurl{URL_TO_YOUR_ARTIFACTS}
\newcommand\vldbpagestyle{plain} 
\usepackage{lipsum}
\usepackage{algorithm}
\usepackage{algpseudocodex}
\usepackage[skip=4pt,font=small]{subcaption}
\usepackage[flushleft]{threeparttable}
\usepackage{amsmath}
\usepackage{amsfonts}
\usepackage{multirow}
\usepackage{xspace}
\usepackage{enumitem}
\usepackage{bm}
\usepackage{siunitx}
\usepackage{dblfloatfix}
\usepackage{soul}
\usepackage{titlesec}

\graphicspath{{fig/}}
\DeclareGraphicsExtensions{.pdf,.eps,.png,.jpg,.jpeg}
\DeclareMathOperator*{\argmax}{arg\,max}
\DeclareMathOperator*{\argmin}{arg\,min}
\newcommand{\refsec}[1]{Section~\ref{#1}}
\newcommand{\refeq}[1]{Eq.~(\ref{#1})}
\newcommand{\reffig}[1]{Figure~\ref{#1}}
\newcommand{\reftab}[1]{Table~\ref{#1}}
\newcommand{\refalg}[1]{Algorithm~\ref{#1}}

\newcommand{\refdef}[1]{Definition~\ref{#1}}
\newcommand{\reflem}[1]{Lemma~\ref{#1}}
\newcommand{\tname}{\textsc{SCARA}}
\newcommand{\pprname}{\textsc{Feature-Push}}
\newcommand{\rename}{\textsc{Feature-Reuse}}
\newcommand{\basename}{\textsc{Feature-Reuse}}

\newcommand{\blue}[1]{\textcolor{black}{#1}}

\begin{document}

\title{\textsc{SCARA}: Scalable Graph Neural Networks with \\
Feature-Oriented Optimization\vspace{-1.ex}}

\author
[   Ningyi~Liao, 
    Dingheng~Mo, 
    Siqiang~Luo, 
    Xiang~Li, 
    Pengcheng~Yin]
{
    Ningyi~Liao$^{\ast\dagger}$, 
    Dingheng Mo$^{\ast\dagger}$, 
    Siqiang Luo$^\dagger$, 
    Xiang Li$^\ddagger$, 
    Pengcheng Yin$^\mathsection$}
\myauthornote{Both authors contributed equally to this research.}
\affiliation{%
    $^\dagger$Nanyang Technological University, Singapore\,
    $^\ddagger${East China Normal University, China}\,
    $^\mathsection${Google Research, USA}
}

\begin{abstract}
Recent advances in data processing have stimulated the demand for learning graphs of very large scales. Graph Neural Networks (GNNs), being an emerging and powerful approach in solving graph learning tasks, are known to be difficult to scale up. Most scalable models apply node-based techniques in simplifying the expensive graph message-passing propagation procedure of GNN. However, we find such acceleration insufficient when applied to million- or even billion-scale graphs. 
In this work, we propose \tname{}, a scalable GNN with feature-oriented optimization for graph computation. \tname{} efficiently computes graph embedding from node features, and further selects and reuses feature computation results to reduce overhead. Theoretical analysis indicates that our model achieves sub-linear time complexity with a guaranteed precision in propagation process as well as GNN training and inference. We conduct extensive experiments on various datasets to evaluate the efficacy and efficiency of \tname{}. Performance comparison with baselines shows that \tname{} can reach up to $100\times$ graph propagation acceleration than current state-of-the-art methods with fast convergence and comparable accuracy. Most notably, it is efficient to process precomputation on the largest available billion-scale GNN dataset Papers100M (111M nodes, 1.6B edges) in 100 seconds. Our tech report can be found at: {\url{https://sites.google.com/view/scara-techreport}} and source code at: {\url{https://github.com/gdmnl/SCARA-PPR}}. 
\end{abstract}

\maketitle

\pagestyle{\vldbpagestyle}


\titlespacing*{\subsection}  {0pt}{1.2ex plus 1ex minus .2ex}{0.2ex}
\setlength{\textfloatsep}{4.2pt plus 0.8pt minus 3.6pt}
\setlength{\floatsep}{4.2pt plus 0.8pt minus 3.2pt}
\setlength{\intextsep}{4.2pt plus 0.8pt minus 3.8pt}
\setlength{\abovedisplayskip}{2.4pt plus 0.2pt minus 2.0pt}
\setlength{\abovedisplayshortskip}{2.2pt plus 0.2pt minus 2.0pt}
\setlength{\belowdisplayskip}{2.4pt plus 0.2pt minus 2.0pt}
\setlength{\belowdisplayshortskip}{2.2pt plus 0.2pt minus 2.0pt}

\def\istech{1}

\begin{table*}[t]
\caption{Time and memory complexity of scalable GNN models. 
Precomputation memory complexity indicates the usage of intermediate variables, while the training and inference memory refer to the GPU usage for storing and updating representation and weight matrices in each training iteration. 
Time complexity is for the full training and/or inference node set. }
\label{tab:complexity}
\centering
\small
\begin{tabular}{@{}c|lll|lll@{}}
	\toprule
	\multirow{2}{*}{\textbf{Model}} 
	    & \textbf{Precomputation} & \textbf{Training} & \textbf{Inference} 
	    & \textbf{Precomputation} & \textbf{Training} & \textbf{Inference} \\
	    & \textbf{Memory}         & \textbf{Memory}   & \textbf{Memory}    
	    & \textbf{Time}           & \textbf{Time}     & \textbf{Time}      \\
	\hline
    GCN \cite{kipf2016semi} 
        & -- & $O(LnF + LF^2)$ & $O(LnF + LF^2)$ 
        & -- & $O(ILmF + ILnF^2)$ & $O(LmF + LnF^2)$ \\ 
    GraphSAINT \cite{zeng2019graphsaint} 
        & -- & $O(L^2 bF + LF^2)$ & $O(LnF + LF^2)$ 
        & -- & $O(IL^2nF^2)$ & $O(LmF + LnF^2)$ \\ 
    GAS \cite{Fey2021}
        & $O(LnF)$ & $O(LdbF + LF^2)$ & $O(LdbF + LF^2)$ 
        & $O(m + LnF)$ & $O(ILmF + ILnF^2)$ & $O(nF)$ \\
    APPNP \cite{Klicpera2019} 
        & $O(m)$ & $O(LbF + LF^2 + db)$ & $O(LbF + LF^2 + db)$ 
        & $O(m)$ & $O(ITmF + ILnF^2)$ & $O(TmF + LnF^2)$ \\ 
    PPRGo \cite{Bojchevski2020} 
        & $O(n/r_{max})$ & $O(LbF + LF^2 + Kb)$ & $O(LbF + LF^2 + Kb)$ 
        & $O(m/r_{max})$ & $O(IKnF + ILnF^2)$ & $O(KnF + LnF^2)$ \\ 
    SGC \cite{wu19sim} 
        & $O(m)$ & $O(LbF + LF^2)$ & $O(LbF + LF^2)$ 
        & $O(LmF)$ & $O(ILnF^2)$ & $O(LnF^2)$ \\ 
    GBP \cite{Chen2020a} 
        & $O(nF)$ & $O(LbF + LF^2)$ & $O(LbF + LF^2)$ 
        & $O(LF \sqrt{Lm \log(Ln)} /\epsilon)$ & $O(ILnF^2)$ & $O(LnF^2)$ \\ 
    \textbf{\tname{} (ours)}
        & $O(nF)$ & $O(LbF + LF^2)$ & $O(LbF + LF^2)$ 
        & $O(F\sqrt{m \log n} /\lambda)$ & $O(ILnF^2)$ & $O(LnF^2)$ \\
	\bottomrule
\end{tabular}
\vspace{-2.5ex}
\end{table*}

\vspace{-1.ex}

\section{Introduction}
\label{sec:introduction}

Recent years have witnessed the burgeoning of online services based on data represented by graphs, which leads to rapid increase in the amount and complexity of such graph data.
Graph Neural Networks (GNNs) are specialized neural models designed to represent and process graph data, and have 
achieved strong performance on graph understanding tasks such as node classification \cite{hamilton2017inductive,kipf2016semi,chen2018fastgcn,chiang2019cluster}, link prediction \cite{velivckovic2017graph,zhang2018link,berg2017graph,Ying2018graph}, and community detection \cite{wang2017mgae,Rami2019DDGK,Chen2019a}. 

One of the most widely adopted GNN models is the Graph Convolutional Network (GCN) \cite{kipf2016semi} which learns graph representations by leveraging information of topological structure. 
Specifically, the GCN represents each node state by a feature vector, successively propagates the state to neighboring nodes, and updates the neighbor features using a neural network.
This interleaved process of graph propagation and state update can proceed for multiple iterations. 

While being able to effectively gather state information from the graph structure, GCNs are known to be resource-demanding, which implies limited scalability when deployed to large-scale graphs \cite{Wu2021a,Zhu2022}. 
It is also non-trivial to fit the node features of large graphs into the memory of hardware accelerators like GPUs.
However, it is increasingly demanding to apply these effective models to modern real-world graph datasets.
Recent studies have attempted to learn representations of large graphs such as the Microsoft Academic Graph (MAG) of 100 million entries \cite{Sinha2015an,yang2020scaling}. 
Nonetheless, directly fitting the basic GCN model to such data would easily cause unacceptable training time or out-of-memory error. 
Hence, how to adopt the GCN model efficiently to these very large-scale graphs while benefiting from its performance becomes a challenging yet important problem in realistic applications. 

\noindentparagraphb{Existing Approaches are Not Scalable Enough. } 
Several techniques have been proposed towards more efficient learning for GNN, addressing the scalability issues. One optimization is to decouple graph propagation from the feature update and employ linear models to simplify computation \cite{wu19sim,Klicpera2019}. There is no need to store the whole graph in the GPU and the memory footprint is thence reduced. Such methods exploit graph data management techniques such as Personalized PageRank \cite{Page1999} to calculate the graph representation used in the model. 
Another direction is easing node interdependence, which enables training on smaller batches and is achieved by neighbor sampling \cite{Ahmed2017,Chen2018}, layer sampling \cite{chen2018fastgcn,hamilton2017inductive,Fey2021}, and subgraph sampling \cite{chiang2019cluster,zeng2019graphsaint,Huang2021}. Various sampling schemes have been applied to restrain the number of nodes contained in GNN learning pipelines and reduce computational overhead. 
Other algorithms are also utilized in simplifying graph propagation and learning in order to improve efficiency and efficacy, including diffusion \cite{Klicpera2019,Atwood2016}, self-attention \cite{velivckovic2017graph,Thekumparampil2018,zhang2020graph}, and quantization \cite{Ding2021}. 

Unfortunately, such methods are still not efficient enough when applied to million-scale or even larger graphs. 
According to \cite{Wang2021}, the very recent state-of-the-art algorithm GBP \cite{Chen2020a} still consumes more than $10^4$ seconds solely for precomputation on the Papers100M graph ($111\si{M}$ nodes, $1.6\si{B}$ edges, generated from MAG) to reach proper accuracy.
In our experiments, the same model even exceeds the $160\si{GB}$ RAM bound on a single worker during learning. Such cost is still too high for the method to be applied in practice. 




\noindentparagraphb{Our Contributions. }
In this paper, we propose \tname{}, a scalable Graph Neural Network algorithm with low time complexity and high scalability on very large datasets. On the theoretical side, the time complexity of \tname{} for precomputation/training/inference matches the same sub-linear level with the state of the art, as shown in \reftab{tab:complexity}. 
On the practical side, to our knowledge, \tname{} is the first GNN algorithm that can be applied to billion-scale graph Papers100M with a precomputation time less than 100 seconds and complete training under a relatively strict memory limit. 

Particularly, \tname{} employs several feature-oriented optimizations. First, we observe that most current scalable methods repetitively compute the graph propagation information from the node-based dimension, which results in complexity at least proportional to the number of graph nodes. To address this issue, we design a {\pprname{}} method that realizes the information propagation from the feature vectors, which removes the linear dependency on the number of nodes in the complexity while maintaining the same precision of corresponding graph propagation values. 
Second, as we mainly process the feature vectors, we discover that there is significant room to reuse the computation results across different feature dimensions. Hence we propose the \rename{} algorithm. Through compositing the calculation results, \tname{} can efficiently prevent time-consuming repetitive propagation. 
By such means, \tname{} outperforms all leading competitors in our experiments in all 6 GNN learning tasks in regard to model convergence time, i.e., the sum of precomputation and training time, with highly efficient inference speed, significantly better memory overhead, and comparable or better accuracy. 


In summary, we have made the following contributions:
\begin{itemize}[wide,labelwidth=!,labelindent=0pt,itemsep=0pt,topsep=0mm]
    \item \blue{We present the \pprname{} algorithm which propagates the graph information from the feature vectors with forward push and random walk. Our method realizes a sub-linear complexity for precomputation running time along with efficient model training and inference implemented in the mini-batch approach. }
    \item We propose the \rename{} mechanism which further utilizes the feature-oriented optimizations to improve the efficiency of feature propagation while maintaining precision. 
    The technique is able to half the overhead for several graph representations. 
    \item We conduct comprehensive experiments to evaluate the efficiency and effectiveness of the \tname{} model on various datasets and with benchmark methods. Our model is efficient to process the billion-scale dataset Papers100M. It also achieves up to $100\times$ faster in precomputation time than the current state of the art.
\end{itemize}

\noindentparagraph{Paper Organization. }
The rest of the paper is organized as follows. \refsec{sec:preliminary} introduces the preliminaries of GNN, with categories and complexity of scalable methods. Then we propose our \tname{} framework in \refsec{sec:method}, introducing the \pprname{} and \rename{} algorithms. Experimental results are presented in \refsec{sec:experiment}, including evaluations on efficacy, efficiency, and convergence. Finally, we summarize the paper in \refsec{sec:conclusion}. 


\section{Preliminaries and Related Works}
\label{sec:preliminary}
\blue{In this section, we introduce the concept of GNN and review the scalable benchmark models related to our design. }

\noindentparagraph{Notations. }
Consider a graph $G = \langle V, E\rangle$ with node set $V$ and edge set $E$. 
Let $n = |V|$, $m = |E|$, and $d = m/n$, respectively. The graph connectivity is represented by the adjacency matrix with self-loops as $\bm{A} \in \mathbb{R}^{n \times n}$, while the diagonal degree matrix is $\bm{D} \in \mathbb{R}^{n \times n}$. Following \cite{Chen2020a,Wang2021}, we generalize the adjacency matrix of $G$ normalized by $\bm{D}$ with convolution coefficient $r \in [0, 1]$ as $\Tilde{\bm{A}}_{(r)} = \bm{D}^{r-1} \bm{A} \bm{D}^{-r}$. For each node $v \in V$, denote the set of the out-neighbors of $v$ by $\mathcal{N}(v) = \{u | (v, u) \in E \}$, and the out-degree of $v$ by $d(v) = |\mathcal{N}(v)|$. 
Node $v$ is represented by an $F$-dimension attribute vector $\bm{x}(v)$, which composes the attribute matrix $\bm{X} \in \mathbb{R}^{n \times F}$. 

A GNN recurrently computes the node representation matrix $\bm{H}^{(l)}$ as current state in the $l$-th layer to achieve an optimal output prediction. The model input feature matrix is $\bm{H}^{(0)} = \bm{X}$ in particular. 
For a conventional $L$-layer GCN \cite{kipf2016semi}, the $(l+1)$-th representation matrix $\bm{H}^{(l+1)}$ is updated as: 
\begin{equation}
    \bm{H}^{(l+1)} = \sigma \left( \Tilde{\bm{A}} \bm{H}^{(l)} \bm{W}^{(l)} \right), \quad l = 0, 1, \cdots, L-1,
    \label{eq:gcn}
\end{equation} 
\noindent where $\bm{W}^{(l)}$ is the trainable weight matrix of the $l$-th layer, $\Tilde{\bm{A}} = \Tilde{\bm{A}}_{(1/2)}$ is the normalized adjacency matrix, and $\sigma (\cdot)$ is the activation function such as ReLU or softmax. For analysis simplicity we keep the feature size $F$ unchanged in all layers. 

We here present an analysis on the complexity bounds of the GCN model in \refeq{eq:gcn} to explain the restraints of its efficiency. 
\blue{We mainly focus on the training process as it requires most resources for traversing whole graph and updating the model for a large number of $I$ epochs. }
For the $L$-layer GCN model, the multiplication of \textit{graph propagation} $\Tilde{\bm{A}} \bm{H}^{(l)}$ is bounded by a complexity of $O(L m F)$ giving the adjacency matrix $\Tilde{\bm{A}}$ with $m$ entries and the propagation is conducted for $L$ iterations. The overhead for \textit{feature transformatio}n by multiplying $\bm{W}^{(l)}$ is $O(L n F^2)$. 
In the \textit{training} stage, the above procedure is repeated to iteratively update the model weights $\bm{W}^{(l)}$. 
As discovered by previous studies \cite{chiang2019cluster,Chen2020a}, the dominating term is $O(L m F)$ when the graph is large, since the latter transformation can be accelerated by parallel computation. Hence, the full graph propagation becomes the scalability bottleneck. 
For memory usage, the GCN typically takes $O(LnF + LF^2)$ space to store the node representation and weight matrices in its layers. 

\noindentparagraph{Post-Propagation Model. }
As the graph propagation possesses the major computation overhead when the graph is scaled-up, a straightforward idea is to simplify this step and prevent it from being repetitively included in each layer. Such approaches are regarded as propagation decoupling models \cite{zhang2020,Li2022}. We further classify them into post- and pre-propagation variants based on the presence stage of propagation relative to feature transformation. 

\blue{The post-propagation decoupling methods apply propagation only on the last model layer}, enabling efficient and individual computation of the graph propagation matrix, as well as the fast and simple model training. 
The APPNP model \cite{Klicpera2019} introduces the personalized PageRank (PPR) \cite{Page1999} algorithm in the propagation stage. The iterative graph propagation in the GCN updates is replaced by multiplying the PPR matrix after the feature transformation layers:
\begin{align}
    \bm{H}^{(l+1)} &= \sigma \left( \bm{H}^{(l)} \bm{W}^{(l)} \right), \quad l = 0, 1, \cdots, L-2, \label{eq:post_1} \\
    \bm{H}^{(l+1)} &= \sigma \left( \hat{\bm{\Pi}} \bm{H}^{(l)} \bm{W}^{(l)} \right), \quad l = L-1, \label{eq:post_2}
\end{align}
\noindent where $\hat{\bm{\Pi}} = \sum_{l=0}^L \alpha (1-\alpha)^l \Tilde{\bm{A}}^l$ is the approximate PPR matrix with teleport probability $\alpha$. 

In such design, the feature transformation benefit from mini-batch scheme in both training and inference stage, hence reduces the demand of GPU memory. In \reftab{tab:complexity}, the batch size is $b$. Regarding computation speed, a $T$-round Power Iteration computation on the PPR matrix \cite{Page1999} leads to $O(TmF + LnF^2)$ time per epoch. 
The PPRGo model \cite{Bojchevski2020} further improves the efficiency of precomputing the PPR matrix $\bm{\Pi}$ by the Forward Push algorithm \cite{Andersen2006} with error threshold $r_{max}$ and only records the top-$K$ entries. However, it demands $O(n/r_{max})$ space to store the dense PPR matrix. 

\noindentparagraph{Pre-Propagation Model. }
Another line of research, namely the pre-propagation models such as SGC \cite{wu19sim}, chooses to propagate graph information in advance and encode to the attributes matrix $\bm{X}$, forming an embedding matrix $\bm{P}$ which is utilized as the input feature to the neural network layers. In a nutshell, we summarize the model updates in the following scheme: 
\begin{align}
    \bm{H}^{(0)}   &= \bm{P} = \sum_{l=0}^{L_P} a_l \Tilde{\bm{A}}_{(r)}^l \cdot \bm{X}, \label{eq:overview_pre}\\
    \bm{H}^{(l+1)} &= \sigma \left( \bm{H}^{(l)} \bm{W}^{(l)} \right), \quad l = 0, 1, \cdots, L-1, \label{eq:overview_trans}
\end{align}
\noindent where $L_P$ denotes the depth of precomputed propagation and $a_l$ is the layer-dependent diffusion weight. 

The line of \refeq{eq:overview_pre} corresponds to the \textit{precomputation section} and is calculated only once for each graph. 
The complexity of this stage is solely related with the precomputation techniques applied in the model. In SGC, the equation is given by an $L$-hop multiplication of $\bm{P} = \Tilde{\bm{A}}^L \bm{X}$, taking $O(LmF)$ time. 
A recent work GBP \cite{Chen2020a} employs a PPR-based bidirectional propagation with $L_P = L$ and tunable $a_l$ and $r$. Under an approximation of relative error $\epsilon$, it improves precomputation complexity to $O(LF \sqrt{Lm \log(Ln)}/\epsilon)$ in the best case. 
\blue{It is notable that since GBP contains a node-based traverse scheme, it is sensitive to the scale of $n$ in practice. }

\refeq{eq:overview_trans} follows the neural network \textit{feature transformation}, taking $\bm{P}$ as input feature. 
\blue{Comparing to \refeq{eq:post_2}, it completely removes the need of additional multiplication, hence both training and inference is reduced to $O(LnF^2)$. 
The simple GNN provides scalability in both resource-demanding training and frequently-queried inference, with the ease to employ techniques such as mini-batch training, parallel computation, and data augmentation. }

\noindentparagraph{Other Methods. }
\blue{There is a large scope of GNNs related to sampling techniques, which simplifies the propagation by replacing the full-batch graph updates with sampled nodes in mini-batches. 
A popular direction is graph-wise sampling, such as Cluster-GCN \cite{chiang2019cluster} exploiting clustering structure and GraphSAINT \cite{zeng2019graphsaint} using various levels of information. }
The representative GraphSAINT-RW integrates $L$-hop random walk on samples and produces a training complexity of $O(L^2nF + LnF^2)$. 
The sampling are not applicable in the full graph inference stage, causing the inference time and memory overheads being the same with the vanilla GCN. 
\blue{GAS \cite{Fey2021} samples layer-wise neighbors to achieve pruned propagation and spends large memory overhead for historical embedding to save computation. Its $O(LmF + LnF^2)$ training complexity is counted as each edge is traversed once per layer, while the optimal inference complexity is benefited by the cache embedding. }

In our experiments, we compare our GNN algorithm with the state of the arts from each of the aforementioned categories, to demonstrate the scalability and effectiveness of our algorithm.


\section{\tname{} Framework}
\label{sec:method}
\blue{We propose our \tname{} framework, which composes two major optimizations, namely {\pprname} and {\rename}. The \pprname{} algorithm conducts propagation from the aspect of feature, while {\rename} is a novel technique that reuses columns in the feature matrix. 
We present thorough analysis on the algorithmic complexity and precision guarantee to demonstrate the theoretical validity and effectiveness of \tname{}. }

\subsection{Overview}
\label{ssec:overview}
To realize scalability in the network training and inference stage, and to better employ advanced Personalized PageRank (PPR) algorithms to optimize graph diffusion, we apply the backbone of propagation decoupling approach in our GNN design. Similar to the idea of pre-propagation models \cite{wu19sim,Chen2020a}, in precomputation stage we follow \refeq{eq:overview_pre} to compute the graph information $\bm{P}$ in advance together with the node attributes $\bm{X}$. Then, a simple yet effective feature transformation is conducted as given in \refeq{eq:overview_trans}. 

The propagation stage is the complexity bottleneck, mentioned earlier. Hence, we focus on reducing its computation complexity. We rewrite \refeq{eq:overview_pre} in our propagation as:
\begin{equation}
    \bm{P} = \sum_{l=0}^{\infty} \alpha (1-\alpha)^l \Tilde{\bm{A}}_{(r)}^l \cdot \bm{X}
    = \sum_{l=0}^{\infty} \alpha (1-\alpha)^l \left( \bm{D}^{r-1} \bm{A} \bm{D}^{-r} \right)^l \bm{X}, \label{eq:overview_p}
\end{equation}
\noindent where $\alpha$ is the teleport probability as we set $a_l = \alpha (1-\alpha)^l$ to be associated with the form in the PPR calculation. Compared with APPNP and PPRGo, we adopt a generalized graph adjacency $\Tilde{\bm{A}}_{(r)}$ with an adjustable convolution factor $r \in [0, 1]$ to fit different scales of graphs. The upper bound is set to $L_P = \infty$ to better capture the whole graph information without efficiency degeneration. 

Our computation of \refeq{eq:overview_p} is displayed in \refalg{alg:featpush} (\pprname{}) and explained in detail in \refsec{ssec:fwdpush}. The highlight of \pprname{} is the application of propagating from features, \blue{which differs from prior works}. In many real-world tasks, when a graph being scaled-up, its numbers of nodes ($n$) and edges ($m$) increase, but the node attributes dimension ($F$) usually remains unchanged. Thus, an algorithm with complexity mainly dependent on $F$ enjoys better scalability than those dominated by $n$ or $m$. 

As the attribute matrix $\bm{X}$ is included in our computation, we then investigate how to fully utilize its information contained to further accelerate our algorithm, which leads to the \refalg{alg:featreuse} (\rename{}). The motivation is to reduce the expensive iterative computation of $\bm{P}$ components by exploiting the previous results based on attribute vectors $\bm{x}$ on selected dimensions $f$. \blue{We apply linear combination with precision guarantee to reduce the precision constraint of \refalg{alg:featpush} with improved speed.} We further describe this methodology in \refsec{ssec:freuse}. 

{\small 
\begin{algorithm}[bp]
    \algrenewcommand{\alglinenumber}[1]{\scriptsize\bfseries#1}
    \fontsize{9pt}{12pt}\selectfont
	\renewcommand{\algorithmicrequire}{\textbf{Input:}}
	\renewcommand{\algorithmicensure}{\textbf{Output:}}
	\caption{\pprname{}}
	\label{alg:featpush}
	\begin{algorithmic}[1]
		\Require Graph $G$, node set $U$, feature vector $\bm{x}$,
		    probability $\alpha$, convolution factor $r$, push coefficient $\beta$
		\Ensure Approximate embedding vector $\hat{\bm{\pi}} (\bm{x})$
        \ForAll{$u \in U$}
            \State $r'(\bm{x}; u) \gets x(u) \cdot d(u)^{1-r}$        \label{alg:fp_init_r}
            \State $r(\bm{x}; u) \gets r'(\bm{x}; u) / \sum_{u \in U} r'(\bm{x}; u)$       \label{alg:fp_init_r_norm}
        \EndFor
		\State $\hat{\pi}(\bm{x}; t) \gets 0$ for all $t \in U$                  \label{alg:fp_init_pi}
		\While{exist $u \in U$ such that $r(\bm{x}; u) > r_{max} / d(u)$}
		    \ForAll{$v \in \mathcal{N}(u)$}
		        \State $r(\bm{x}; v) \gets r(\bm{x}; v) + (1-\alpha) \cdot r(\bm{x}; u) / d(u)$    \label{alg:fp_fwd}
		    \EndFor
		    \State $\hat{\pi}(\bm{x}; u) \gets \hat{\pi}(\bm{x}; u) + \alpha \cdot r(\bm{x}; u)$           \label{alg:fp_push}
		    \State $r(\bm{x}; u) \gets 0$        \label{alg:fp_set_r}
		\EndWhile                           \label{alg:fp_end_fwdpush}
        \State $r_{sum} \gets \sum_{u\in U} r(\bm{x}; u)$, \; $N_W \gets r_{sum} / \beta$   \label{alg:fp_w}
		\ForAll{$u \in U$ such that $r(\bm{x}; u) \neq 0$}
		    \State perform $\frac{r(\bm{x}; u)}{r_{sum}} \cdot N_W$ random walks from $u$          \label{alg:fp_rw}
		    \ForAll{random walk stopping at $t$}
		        \State $\hat{\pi}(\bm{x}; t) \gets \hat{\pi}(\bm{x}; t) + r_{sum} / N_W$                      \label{alg:fp_add}
		    \EndFor
		\EndFor                             \label{alg:fp_end_rw}
		\State $\hat{\pi}(\bm{x}; t) \gets \hat{\pi}(\bm{x}; t) \cdot d(t)^{r-1}$ for all $t \in U$        \label{alg:fp_norm}
		\State \Return {$\hat{\bm{\pi}} (\bm{x}) \gets \left( \hat{\pi}(\bm{x}; t_1), \cdots, \hat{\pi}(\bm{x}; s_{n_U}) \right)$}
	\end{algorithmic}
\end{algorithm}
}

\subsection{\pprname{}}
\label{ssec:fwdpush}
Examining \refeq{eq:overview_p}, the embedding matrix $\bm{P}$ is the composition of graph diffusion matrix $\Tilde{\bm{A}}_{(r)}$ and node attributes $\bm{X}$. 
Most scalable methods such as APPNP \cite{Klicpera2019} and SGC \cite{wu19sim} compute the propagation part separately from network training, resulting in a complexity at least proportional to $m$. 
GBP \cite{Chen2020a} discusses a bidirectional propagation with both node-side random walk on $\bm{D}^{-1} \bm{A}$ and feature-side reverse push on $\bm{D}^{-r} \bm{X}$. Although the random walk step ensures precision guarantee, it requires long running time when not being accelerated by other methods \cite{Wang2017,Wang2019}. 

We propose the \pprname{} approach that propagates graph information from the feature dimension, which is capable to utilize efficient single-source PPR algorithms through a simple but surprisingly effective transformation. Intuitively, the \pprname{} is first initialized by the normalized features $\bm{D}^{1-r} \bm{X}$. Then, single-source PPR algorithms, which compute the PPR values from a source node to other nodes, are applied to iteratively propagate the information with $\left( \bm{A} \bm{D}^{-1} \right)^l$. It achieves the embedding matrix based on that:
\begin{equation}
 \Tilde{\bm{A}}_{(r)}^l \cdot \bm{X} = \left( \bm{D}^{r-1} \bm{A} \bm{D}^{-r} \right)^l \bm{X} = \bm{D}^{r-1} \left( \bm{A} \bm{D}^{-1} \right)^l \bm{D}^{1-r} \bm{X} ,
 \label{eq:fwdpush_norm}
\end{equation}

\blue{In order to better derive \pprname{}, we borrow the Personalized PageRank (PPR) notations to describe our technique manipulating feature vectors. }
On a graph $G$, given a source node $s \in V$ and a target node $t \in V$, the PPR $\pi(s, t)$ represents the probability of a random walk with teleport factor $\alpha \in (0, 1)$ which starts at node $s$ stops at $t$. In general, \textit{forward} PPR algorithms, often categorized as \textit{single-source} PPR, start the computation from $s$, comparing to \textit{backward} or \textit{reverse} alternatives that are developed from $t$ \cite{Wang2021}. 

When the PPR calculation is integrated with features, it shares similarities in forms but is with different interpretation. 
Consider the PPR problem with nodes in a set $U \subseteq V$ as the source nodes. Let $n_U$ be the size of set $U$. Denote a matrix $\bm{X} = \left(\bm{x}_1, \cdots, \bm{x}_F \right)$, where $\bm{x}_f$ ($1 \le f \le F$) is the $f$-th column vector that is of length $n_U$ and sum of elements is $1$. Following \cite{Wang2021}, we assume all the entries $x_f(u) \ge 0$ for each $u \in U$. 
We use $\pi(\bm{x}; t)$ to represent the PPR for feature vector $\bm{x}$, and can be defined as the probability of the event that a random walk which starts at a node $s \in U$ with probability distribution ${\bm{x}}$ and stops at $t$. 
\blue{It can be derived from the definition that, each feature PPR $\pi(\bm{x}; t)$ can be interpreted as a generalized integration of normal PPR value $\pi(s, t)$, hence the properties and operations of common PPR are still valid. }
The embedding matrix is $\bm{P} = \left(\bm{\pi}(\bm{x}_1), \cdots, \bm{\pi}(\bm{x}_F) \right)$, where $\bm{\pi}(\bm{x}_f) = \bm{\pi}_f$ is the $f$-th column of feature PPR vector corresponding to vector $\bm{x}_f$, and is composed by $\bm{\pi}_f = \left( \pi(\bm{x}_f; t_1), \cdots, \pi(\bm{x}_f; t_{n_U}) \right)$. 
\blue{We here look into the redefined problem for approximating feature PPR:}

\begin{definition}[\textbf{Approximate PPR for Feature}]
\label{thm:featppr}
    Given an absolute error bound $\lambda>0$, a PPR threshold $0<\delta<1$, and a failure probability $0<\phi<1$, the approximate PPR query for feature vector $\bm{x}$ computes an estimation $\hat{\pi}(\bm{x}; t)$ for each $t \in U$ with $\pi(\bm{x}; t) > \delta$, such that with probability at least $1 - \phi$, 
    \begin{equation}
        |\pi(\bm{x}; t) - \hat{\pi}(\bm{x}; t)| \le \lambda.
        \label{eq:fwdpush_appr}
    \end{equation}
\end{definition}

Recognizing that GNNs require less precise propagation information to achieve proper performance \cite{daniel2018adversarial,Sun2018a}, the approximate feature PPR enables employing efficient computation based on forward PPR algorithms without loss in eventual model effectiveness \cite{Wang2017,Wu2021}. 
\blue{We employ a scalable algorithm \pprname{} to compute the embedding matrix combining Forward Push \cite{Andersen2006} and Random Walk techniques that both operate the feature vectors.} The algorithm makes use of both approaches, that random walk is accurate but less efficient, while forward push is fast with a loose precision guarantee. 
Algorithms exploiting such combination have been the state of the arts in various PPR benchmarks \cite{Wang2017,Lin2020index}. \blue{We highlight that the differences between our \refalg{alg:featpush} and the algorithms in \cite{Wang2017,Lin2020index} are two-fold. First, the push starts from the feature vector, which can be seen as a generalized PPR operation taking probability distribution $\bm{x}$ into account. Second, the repetitive queries on different features offer the ease for transformation in \refeq{eq:fwdpush_norm} and optimization in \refsec{ssec:freuse}. }

As shown in \refalg{alg:featpush}, the \pprname{} algorithm outputs the approximation of embedding vector $\hat{\bm{\pi}}(\bm{x})$ for input feature $\bm{x}$. 
Repeating it for $F$ times with all features $\bm{x}_1, \cdots, \bm{x}_F$ produces all columns compositing the estimate of embedding matrix $\hat{\bm{P}}$. 
The algorithm first computes the approximation $\hat{\pi}(\bm{x}; t)$ for each node $t \in U$ through forward push (line \ref{alg:fp_init_r}-\ref{alg:fp_end_fwdpush} in \refalg{alg:featpush}), then conducts compressed random walks to save computation (line \ref{alg:fp_w}-\ref{alg:fp_end_rw}). We analyze each methods and their combination respectively. 

\noindentparagraph{\blue{Forward Push on Feature Value.} }
\blue{Instead of calculating the PPR value $\pi(s, t)$, the forward push method in \pprname{} maintains a \textit{reserve value} $\hat{\pi}(\bm{x}; t)$ directly for node $t \in U$ and feature $\bm{x}$ as the estimation of $\pi(\bm{x}; t)$.} An auxiliary \textit{residue value} $r(\bm{x}; t)$ is recorded as the intermediate result for each node-feature pair. 
\blue{To transfer node attributes to distributions in line with $\pi(\bm{x}; t)$ that stands for the probability with a sum of $1$ for all nodes $t \in U$, the $L_1$-normalized feature vector $\bm{x}$ are used to initialize the residues. }
The forward push algorithm subsequently updates the residue of target node $t$ from the source node $s$ to propagate the information. The threshold $r_{max}$ controls the terminating condition so that the process can stop early. Eventually, the forward push transfers $\alpha$ portion of node residue $r(\bm{x}; t)$ into reserve value, while distributing the remaining $(1-\alpha)$ to the neighbors of $s$. 

\noindentparagraph{Random Walk on Feature Residue. } 
\pprname{} then performs random walks with decay factor $\alpha$ to propagate the residue feature value. 
Compared with the pure random walk approach, \pprname{} only requires $\frac{r(\bm{x}; t)}{r_{sum}} \cdot N_W$ number of walks per node with the same precision guarantee, benefiting from the Forward Push results. 
The estimation of $\hat{\pi}(\bm{x}; t)$ is achieved by implementing the Monte-Carlo method \cite{Fogaras2005,Wang2019}, and is updated according to the fraction of random walks terminating at $t$. 

\noindentparagraph{Combination and Normalization. }
To depict the combination of forward push residual and random walk numbers, we firstly define the push coefficient $\beta$:
\begin{definition}[\textbf{Push Coefficient}]
    The push coefficient $\beta$ is the scale between the total left residual $r_{sum}$ and the total number of sampled random walks $N_W$ in  \pprname{}. 
\label{def:push coef}
\end{definition}

\blue{The scale $\beta$ is the key coefficient of \pprname{}, which balances absolute error guarantee and time complexity. Referencing the trade-off in \cite{Wang2017}, we set $\beta$ to a specific value, namely standard push coefficient $\beta_s = \frac{\lambda^2}{(2\lambda/3+2) \cdot \log(2/\phi)}$, to satisfy the guarantee of $\hat{\pi}(\bm{x}; t)$ in \refdef{thm:featppr}.
In \refalg{alg:featpush}, the forward push and random walk are combined as line \ref{alg:fp_add}.} Derived from the single-source PPR analysis \cite{Andersen2006,Wang2017}, we state that our \pprname{} algorithm provides an unbiased estimation $\hat{\pi}(\bm{x}; t)$ of the value $\pi(\bm{x}; t)$:
\begin{lemma}
\label{thm:g_featpush}
     \refalg{alg:featpush} produces an unbiased estimation  $\hat{\pi}(\bm{x}; t)$ of the value $\pi(\bm{x}; t)$ satisfying \refeq{eq:fwdpush_appr}. Repeating it for $F$ times produces an unbiased estimation $\hat{\bm{P}}$ of the embedding matrix $\bm{P}$. 
\end{lemma}

The forward push and random walk combination produces approximate $\bm{\Pi}^{(l)} = \alpha (1-\alpha)^l \left(\bm{A} \bm{D}^{-1} \right)^l$ for a certain $l$. To be aligned with the embedding matrix $\bm{P}^{(l)}$ in \refeq{eq:overview_p}, we apply the normalization by degree vector (lines \ref{alg:fp_init_r} and \ref{alg:fp_norm} in \refalg{alg:featpush}) to achieve the transformation in \refeq{eq:fwdpush_norm}. 
\blue{These operations on embedding values can be efficiently implemented in vector-based schemes. }

\subsection{\rename{}}
\label{ssec:freuse}

\begin{algorithm}[!b]
    \algrenewcommand{\alglinenumber}[1]{\scriptsize\bfseries#1}
    \fontsize{9pt}{12pt}\selectfont
	\renewcommand{\algorithmicrequire}{\textbf{Input:}}
	\renewcommand{\algorithmicensure}{\textbf{Output:}}
	\caption{\blue{\rename{}}}
	\label{alg:featreuse}
	\begin{algorithmic}[1]
		\Require \blue{Graph $G$, feature set $\bm{X}=\{\bm{x}_{f}\}$, base set size $n_B$, decomposition threshold $\delta_0$, precision factor $\gamma$, error bound $\lambda$}
		\Ensure Approximate embedding matrix $\hat{\bm{P}}$
		\State \blue{$\beta_s \gets \frac{\lambda^2}{(2\lambda/3+2) \cdot \log(2n)}$}
		\State $M(\bm{x}_f) \gets 0$ for all $\bm{x}_f \in \bm{X}$,\; $\bm{X}_B = \emptyset$ \label{alg:fr_init}
		\ForAll{$\bm{x}_f \in \bm{X}$}          \label{alg:fr_mfor}
	        \State $\bm{x}_{f^*} \gets \argmin_{\bm{x}_{f^*}\in \bm{X}}{\Vert \bm{x}_{f^*} - \bm{x}_f \Vert_1}$
	        \State $M(\bm{x}_{f^*}) \gets M(\bm{x}_{f^*}) + 1$  \label{alg:fr_m}
		\EndFor
		\For{$i$ from $1$ to $n_B$}   \label{alg:fr_bfor}
	        \State $\bm{b}_i \gets \argmax_{\bm{b}_i \in \bm{X}}{M(\bm{b}_i)}$
	        \State $\bm{X}_B \gets \bm{X}_B \cup \bm{b}_i$, \; $\bm{X} \gets \bm{X} - \bm{b}_i$               \label{alg:fr_setb}
		\EndFor
		\For{$i$ from $1$ to $n_B$}
		    \State $\hat{\bm{\pi}}_i \gets$ Apply Alg.~\ref{alg:featpush} on $\bm{b}_i$ with $\beta^* = \gamma \beta_s$    \label{alg:fr_b}
		\EndFor     \label{alg:fr_end_b}
		\ForAll{$\bm{x}_f \in \bm{X}$}  \label{alg:fr_xfor}
	        \State $\theta_i \gets 0$ for $i$ from $1$ to $n_B$
	        \State \blue{$\bm{x}' \gets \bm{x}_f$, \; $\delta \gets 1$, \; $\vartheta \gets 1$}
	        \While{\blue{$\vartheta \cdot \delta > \delta_0$}}
		        \State $\bm{b}_i\gets \argmin_{\bm{b}_i\in \bm{X}_B}{\Vert \bm{x}' - \bm{b}_i \Vert_1}$ 
		        \State \blue{$\vartheta \gets \argmin_{\vartheta}{\Vert \bm{x}' - \vartheta \bm{b}_i \Vert_1}$, $\delta \gets \delta/2$ }
		        \State \blue{$\bm{x}' \gets \bm{x}' - \vartheta \bm{b}_i$,\; $\theta_i \gets \theta_i + \vartheta$ }
	        \EndWhile
	        \State $\bm{\pi}^*_f \gets$ Apply Alg.~\ref{alg:featpush} on $\bm{x}'$ with  $ \beta' = \left( 1 - \gamma \sum_{i=1}^{n_B} \theta_i \right) \beta_s$
	        \For{$i$ from $1$ to $n_B$}
	            \State $\bm{\pi}^*_f \gets \bm{\pi}^*_f + \theta_i \cdot \hat{\bm{\pi}}_{i}$ 
	        \EndFor
		\EndFor     \label{alg:fr_end_xfor}
		\State \Return {$\hat{\bm{P}} = \left(\bm{\pi}^*_1, \cdots, \bm{\pi}^*_F \right)$}
	\end{algorithmic}
\end{algorithm}

\ifx\istech\undefined
    \begin{table*}[!b]
\vspace{-.8ex}
  \caption{Dataset statistics and parameters. ``Split'' is the percentage of nodes in training/validation/testing set. ``(i)'' and ``(t)'' stand for \ul{i}nductive and \ul{t}ransductive tasks. ``(m)'' and ``(s)'' stand for \ul{m}ultiple and \ul{s}ingle target classifications. }
  \label{tab:dataset}
  \centering
  \small
  \begin{tabular}{@{}c|rrrrc|ccc@{}} \toprule
    \textbf{Dataset} & \textbf{Nodes} $n$ & \textbf{Edges} $m$
        & \textbf{Features} $F$ & \textbf{Classes} $N_c$ & \textbf{Split} 
        & \textbf{Probability} $\alpha$ & \textbf{Convolution} $r$ & \textbf{Common} \\ \midrule
    PPI \cite{hamilton2017inductive} & $56,944$ & $818,716$ 
        & $50$  & $121$ (m) & $0.79/0.11/0.10$ (i) 
        & 0.3   & 0.0   & 
    \multirow{6}{*}{\begin{tabular}{@{}c@{}}
        $\lambda = 1 \times 10^{-4}$ \\
        $n_B = 0.02 n_U$ \\
        $\gamma = 0.2$ \\
        $\delta_0 = 1/16 $ 
    \end{tabular}} \\
    Yelp \cite{zeng2019graphsaint} & $716,847$ & $6,977,410$ 
        & $300$ & $100$ (m) & $0.75/0.10/0.15$ (i) 
        & 0.9   & 0.3   &  \\
    Reddit \cite{hamilton2017inductive} & $232,965$ & $114,615,892$ 
        & $602$ & $41$ (s)  & $0.01/0.04/0.96$ (t) 
        & 0.5   & 0.5   &  \\
    Amazon \cite{chiang2019cluster} & $2,400,608$ & $123,718,024$
        & $100$ & $47$ (s)  & $0.70/0.15/0.15$ (i)
        & 0.2   & 0.2   &  \\
    MAG \cite{yang2020scaling} & $27,394,820$ & $366,143,207$
        & $200$ & $100$ (m)  & $0.01/0.01/0.99$ (t)
        & 0.5   & 0.5   &  \\
    Papers100M \cite{Hu2020} & $111,059,956$ & $1,615,685,872$ 
        & $128$ & $172$ (s)  & $0.78/0.08/0.14$ (t) 
        & 0.2   & 0.5   &  \\
    \bottomrule
  \end{tabular}
\end{table*}

\fi


One of the key differences between the feature vector PPR and the classic single-source PPR is that, in single-source PPR, queries on different nodes are orthogonal to each other, while in feature vector PPR there is similarity between different features. 
\blue{\refalg{alg:featpush} that queries based on features enables the possibility of exploiting this specific attribute of feature vectors and utilizing computed feature PPR to estimate the PPR of another similar feature vector. }

We propose \basename{} algorithm that speeds up the feature PPR computation by leveraging and reusing the similarity between different feature vectors. 
We select a set of vectors as the base vectors from all features and compute their PPR values by \pprname{}. When querying the PPR value on a non-base feature vector, \basename{} separates a segment of the vector that can be obtained by combining the base vectors, and estimate the PPR value of this segment directly with the PPR value of the base vectors without additional \pprname{} computation overhead. 

As a toy example, if we have the PPR $\bm{\pi}(\bm{b})$ for base feature vector $\bm{b} = (0.5, 0.5)$, and need to compute the PPR for $\bm{x} = (0.4, 0.6)$, we can firstly decompose $\bm{x} = (0.4, 0.4) + (0, 0.2)$. We then acquire the PPR for $(0.4, 0.4)$ directly by $0.8\bm{\pi}(\bm{b})$, and just need to compute the PPR value of the residue $(0, 0.2)$. 
The latter PPR calculation is faster due to the reduced dimension and a loose precision bound. 

\noindentparagraph{Base Selection. }
\refalg{alg:featreuse} shows the pseudo code of \basename{}. To represent the similarity between feature vectors, we design a simple yet effective metric, namely the {minimum L1 distance counter} $M(\cdot)$. \basename{} chooses $n_B \ll n_U$ feature vectors with the highest {minimum L1 distance counter} as the base vectors $\bm{b}_i$ to compose the base set $\bm{X}_B = \{\bm{b}_{1}, \cdots, \bm{b}_{n_B}\}$ (line \ref{alg:fr_init}-\ref{alg:fr_setb}). \pprname{} is then invoked to compute the PPR value $\hat{\bm{\pi}}(\bm{b}_i , \beta^*)$ of the base vectors with push coefficient {\small$\beta^* = \gamma \beta_s = \frac{\gamma\lambda^2}{\log(2/\phi)\cdot(2\lambda/3+2)}$}, where $0< \gamma \le 1$ is a tunable precision factor. 

\noindentparagraph{Residue Calculation. }
\refalg{alg:featreuse} then computes the approximate values of the rest features (line \ref{alg:fr_xfor}-\ref{alg:fr_end_xfor}). Given selected base vectors $\bm{b}_i \in \bm{X}_B$, a feature vector can be written in a linear decomposition with residue as $\bm{x}_f = \sum_{i=1}^{n_B} \theta_i \cdot \bm{b}_i + \bm{x}'$, where $0 \le \theta_i < 1$ and $\bm{x}'$ is the residue feature vector. 
We compute the PPR vector $\hat{\bm{\pi}}(\bm{x}', \beta')$ according to the remaining part left in linear decomposition $\bm{x}'$ by \pprname{} with a less precise push coefficient ${\small \beta' = \left( 1 - \gamma \sum_{i=1}^{n_B} \theta_i \right) \beta_s }$. 
Finally, we constitute the estimation as:
\begin{equation}
    \bm{\pi}^*(\bm{x}_{f}) = \sum_{i=1}^{n_B} \theta_i \cdot \hat{\bm{\pi}}(\bm{b}_i, \beta^*) + \hat{\bm{\pi}}(\bm{x}', \beta'). 
    \label{eq:pi_star}
\end{equation}

The PPR of base vectors $\hat{\bm{\pi}}(\bm{b}_i, \beta^*)$ acquired by \pprname{} has its own accuracy guarantee as stated in \reflem{thm:g_featpush}. However, how to assure the other vectors composed by \refeq{eq:pi_star} satisfy the approximation in \refdef{thm:featppr}? To investigate the estimation error of feature vectors, we propose the following lemma. 
\ifx\istech\undefined
    All the missing proofs in this paper can be found in~\cite{techreport}.
    \begin{lemma}
        Given a feature vector $\bm{x}_f$, the ground truth of PPR vector is $\bm{\pi}(\bm{x}_{f})$, and the estimation output by \refeq{eq:pi_star} is ${\bm{\pi}^*}(\bm{x}_{f})$. For any respective element $\pi(\bm{x}_{f}; t)$ and ${\pi}^*(\bm{x}_{f}; t)$,
        $|\pi(\bm{x}_{f}; t) - {\pi}^*(\bm{x}_{f}; t)| \le \lambda$
    holds with probability at least $1 - \phi$, for $\beta'$ such that $\beta'>\beta^*$ and 
    \begin{equation}
        \beta' \leq \frac{\lambda^2/\log(2/\phi)-2\sum_{i=1}^{n_B}\theta_i\beta^*}{2\lambda/3+2}. 
        \label{eq:base_push_pre}
    \end{equation}
    \label{thm:base push}
    \end{lemma}
\else
    \begin{lemma}
        Given a feature vector $\bm{x}_f$, the ground truth of PPR vector is $\bm{\pi}(\bm{x}_{f})$, and the estimation output by \refeq{eq:pi_star} is ${\bm{\pi}^*}(\bm{x}_{f})$. For any respective element $\pi(\bm{x}_{f}; t)$ and ${\pi}^*(\bm{x}_{f}; t)$,
        $|\pi(\bm{x}_{f}; t) - {\pi}^*(\bm{x}_{f}; t)| \le \lambda$
    holds with probability at least $1 - \phi$, for $\beta'$ such that $\beta'>\beta^*$ and 
    \begin{equation}
        \beta' \leq \frac{\lambda^2/\log(2/\phi)-2\sum_{i=1}^{n_B}\theta_i\beta^*}{2\lambda/3+2}. 
        \label{eq:base_push_pre}
    \end{equation}
    \label{thm:base push}
    \end{lemma}

\begin{proof}
According to \cite{Page1999}, PPR for feature can also be interpreted as the solution of following linear system:
\[
\bm{\pi}(\bm{x})=\alpha \bm{x} + (1-\alpha)\bm{A}\bm{D^{-1}}{\pi}(\bm{x})
\]
With above equation, we can further derive that
\[
(\bm{I}-(1-\alpha)\bm{A}\bm{D^{-1}})\bm{\pi}(\bm{x})=\alpha \bm{x} 
\]
Let $\bm{H}=\bm{I}-(1-\alpha)\bm{A}\bm{D^{-1}}$, we have that
\[
\bm{\pi}(\bm{x})=\alpha \bm{H^{-1}} \bm{x} 
\]
With above equation, we can observe that PPR for vectors satisfies the associative law, which means that 
\[
\theta\bm{\pi}(\bm{x})=\bm{\pi}(\theta\bm{x})\ ,\ \  \bm{\pi}(\bm{x}_1)+\bm{\pi}(\bm{x}_2) =\bm{\pi}(\bm{x}_1+\bm{x}_2)
\]
According to the associative law we have
\begin{align*}
    \mathbb{E}[\bm{\pi}^*(\bm{x})]
    =& \sum_{i=0}^{n_B}\theta_i\cdot\mathbb{E}[\hat{\pi}(\bm{b}_i, \beta^*)]+\mathbb{E}[\hat{\pi}(\bm{x}', \beta')]\\
    =& \sum_{i=0}^{n_B}\theta_i{\bm\pi}(\bm{b}_i)+\hat{\bm\pi}(\bm{x}')\\
    =& \bm\pi(\sum_{i=0}^{n_B}\theta_i \bm{b}_i+\bm{x}')
    =\bm\pi(\bm{x})
\end{align*}
Therefore $\bm{\pi}^*(\bm{x})$ is an unbiased estimation of $\bm\pi(\bm{x})$.
For each base, we have already computed $\hat{\pi}(\bm{b}_i, \beta^*)$ with \refalg{alg:featpush}. For the computed result of each base $\bm{b}_i$, assume in the process of \refalg{alg:featpush}, the left residue on each node $v$ before sampling random walks is $r(\bm{b}_i, v)$, the total left residue is $r_{sum}(\bm{b}_i)$, and $N_W(\bm{b}_i) = r_{sum}(\bm{b}_i) / \beta^*$ is the number of random walks sampled. 

As each base is computed independently, adding $\sum_{i=0}^{n_B}\theta_i\hat{\pi}(\bm{b}_i, \beta^*)$ is equivalent to push the feature vector $\theta_i \bm{b}_i$ with the same pattern of the computing process of $\hat{\pi}(\bm{b}_i, \beta^*)$, and then sample $N_W(\bm{b}_i)$ random walks on the remaining residues of at total of $\theta_i r_{sum}(\bm{b}_i)$.

For the computing process of $\theta_i \bm{b}_i$, let us consider the $N_W(\bm{b}_i)$ random walks it generate from all nodes. Let the random variable $X_j(\bm{b}_i, t)$ be value 1 if the $j$-th random walk terminates at $t$, and otherwise be 0. Then we have 
\begin{equation}
    \mathbb{E}\left[\sum_{j=0}^{N_W(\bm{b}_i)}\frac{r_{sum}(\bm{b}_i)}{N_W(\bm{b}_i)}X_j(\bm{b}_i, t)\right] = \sum_{v\in V} r(\bm{b}_i, v) \cdot\pi(v,t)
    \label{eq:expection}
\end{equation}
Consider all base vectors, 
\[
    \mathbb{E}\left[\sum_{i=0}^{n_B}\sum_{j=0}^{N_W(\bm{b}_i)}{\frac{\theta_i r_{sum}(\bm{b}_i)}{N_W(\bm{b}_i)}X_j(\bm{b}_i, t)}\right] = \sum_{i=0}^{n_B}{\sum_{v\in V}\theta_i r(\bm{b}_i, v) \cdot\pi(v,t)}
\]
As we have
\begin{align*}
{\pi}^*(\bm{x}, t) 
=& \sum_{i=0}^{n_B}{\theta_i R(\bm{b}_i,t)} + R(\bm{x}',t)\\ 
+& \sum_{i=0}^{n_B}\sum_{j=0}^{N_W(\bm{b}_i)}{\frac{\theta_i r_{sum}(\bm{b}_i)}{N_W(\bm{b}_i)}X_j(\bm{b}_i, t)} 
+ {\frac{r_{sum}(\bm{x}')}{N_W(\bm{x}')}X_j(\bm{x}', t)}
\end{align*}
According to Lemma 3.2 in \cite{Wang2019}, we further get that
\begin{equation}
Pr[|{\pi}^*(\bm{x}, t) - {\pi}(\bm{x}, t)| > \lambda] \leq 2\cdot exp(-\frac{\lambda^2 N_{sum}}{2\nu+2a\lambda/3})
\label{equal:fora lemma}
\end{equation}
where $N_{sum}=N_W(\bm{x}')+\sum_{i=0}^{n_B}N_W(\bm{b}_i)$, 
$a=max\{\frac{\theta_1 r_{sum}(b_1)N_{sum}}{N_W(b_1)}, \cdots,$\\ $ \frac{\theta_{n_B} r_{sum}(b_{n_B})N_{sum}}{N_W(b_{n_B})}, \frac{ r_{sum}(\bm{x}')N_{sum}}{N_W(\bm{x}')}\}$, and 
\begin{align}
\nu =& \frac{1}{N_{sum}}\sum_{i=0}^{n_B}\sum_{j=0}^{N_W(\bm{b}_i)}{\left(\frac{\theta_i r_{sum}(\bm{b}_i)N_{sum}}{N_W(\bm{b}_i)}\right)^2 \mathbb{E}[X_j(\bm{b}_i, t)]}\nonumber\\
+& \frac{1}{N_{sum}}\sum_{j=0}^{N_W(\bm{x}')}{\left(\frac{ r_{sum}(\bm{x}')N_{sum}}{N_W(\bm{x}')}\right)^2 \mathbb{E}[X_j(\bm{x}', t)]}
\label{eq:v}
\end{align}

Recall that $\beta'<\beta^*$ 
, therefore $\frac{ r_{sum}(\bm{x}')N_{sum}}{N_W(\bm{x}')} > \frac{ r_{sum}(\bm{b_i})N_{sum}}{N_W(\bm{b_i})}$ holds for any $\bm{b_i}$. Applying this to $a$, we have that $a = \frac{ r_{sum}(\bm{x}')N_{sum}}{N_W(\bm{x}')}$. 

To simplify \refeq{eq:v}, we rewrite it with \refeq{eq:expection} as
\begin{align}
\nu=&\frac{1}{N_{sum}}\sum_{i=0}^{n_B}\frac{\theta_i^2 r_{sum}(\bm{b}_i)N_{sum}^2}{N_W(\bm{b}_i)}\cdot{\sum_{v\in V}r(\bm{b}_i, v) \cdot\pi(v,t)}\nonumber\\
+&\frac{1}{N_{sum}^2}\frac{ r_{sum}(\bm{x}')N_{sum}}{N_W(\bm{x}')}\cdot{\sum_{v\in V}r(\bm{x'}, v) \cdot\pi(v,t)}\\
\leq&\sum_{i=0}^{n_B}\frac{\theta_i^2 r_{sum}(\bm{b}_i)N_{sum}^2}{N_W(\bm{b}_i)} +  \frac{ r_{sum}(\bm{x}')N_{sum}}{N_W(\bm{x}')}
\end{align}
Recall that by Definition \ref{def:push coef}, the coefficients $\beta^* = r_{sum}(\bm{b}_i)/N_W(\bm{b}_i)$, $\beta' = r_{sum}(\bm{x}')/N_W(\bm{x}')$, we derive \refeq{equal:fora lemma} as:
\[
Pr[|{\pi}^*(\bm{x}, t) - {\pi}(\bm{x}, t)| > \lambda] \leq 2\cdot exp(-\frac{\lambda^2}{2\sum_{i=0}^{n_B}\theta_i^2\beta^*+2\beta'+2\beta'\lambda/3})
\]
By setting $\beta'$ as
\[
 \beta' \leq \frac{\lambda^2/\log(2/\phi)-2\sum_{i=0}^{n_B}\beta^*\theta_i}{2\lambda/3+2}
 \]
It hence has 
\[
Pr[|{\pi}^*(\bm{x}, t) - {\pi}(\bm{x}, t)| > \lambda] \leq \phi
\]

\end{proof}

\fi

\vspace{-1ex}
\reflem{thm:base push} indicates that, when choosing a smaller push coefficient $\beta^*$ for base vectors, the coefficient $\beta'$ can be larger and reduce the cost of PPR computation on most feature vectors.
We can thence derive the following lemma, which states that the setting in \refalg{alg:featreuse} satisfies \refdef{thm:featppr}:
\begin{lemma}
    Given a feature vector set $\bm{X}$, for any feature vector $\bm{x}_f\in \bm{X}$, \refalg{alg:featreuse} returns an approximate PPR vector $\hat{\bm{\pi}}(\bm{x}_f)$, that any of its element $\hat{\pi}(\bm{x}_f; t)$ satisfies \refeq{eq:fwdpush_appr} with at least $1-\phi$ probability.
\end{lemma}

\begin{proof}
    In \refalg{alg:featreuse} there is $\beta^* = \gamma\beta_s$. 
    Then the push coefficient $\beta' = \left( 1 - \gamma \sum_{i=1}^{n_B} \theta_i \right) \beta_s$ in \refalg{alg:featreuse} satisfies:
\begin{equation}
  \small
  \beta' 
    \leq \frac{\lambda^2/\log(2/\phi)}{2\lambda/3+2}-\sum_{i=1}^{n_B}\beta^*\theta_i\\
    \leq \frac{\lambda^2/\log(2/\phi)-2\sum_{i=1}^{n_B}\beta^*\theta_i}{2\lambda/3+2}.
\end{equation}

Therefore, $\beta^*$ for base vectors and $\beta'$ for remaining vectors satisfy \refeq{eq:base_push_pre}. According to \reflem{thm:base push} this lemma follows.
\end{proof}

\ifx\istech\undefined
\else
    \begin{table*}[!b]
\vspace{-.8ex}
  \caption{Dataset statistics and parameters. ``Split'' is the percentage of nodes in training/validation/testing set. ``(i)'' and ``(t)'' stand for \ul{i}nductive and \ul{t}ransductive tasks. ``(m)'' and ``(s)'' stand for \ul{m}ultiple and \ul{s}ingle target classifications. }
  \label{tab:dataset}
  \centering
  \small
  \begin{tabular}{@{}c|rrrrc|ccc@{}} \toprule
    \textbf{Dataset} & \textbf{Nodes} $n$ & \textbf{Edges} $m$
        & \textbf{Features} $F$ & \textbf{Classes} $N_c$ & \textbf{Split} 
        & \textbf{Probability} $\alpha$ & \textbf{Convolution} $r$ & \textbf{Common} \\ \midrule
    PPI \cite{hamilton2017inductive} & $56,944$ & $818,716$ 
        & $50$  & $121$ (m) & $0.79/0.11/0.10$ (i) 
        & 0.3   & 0.0   & 
    \multirow{6}{*}{\begin{tabular}{@{}c@{}}
        $\lambda = 1 \times 10^{-4}$ \\
        $n_B = 0.02 n_U$ \\
        $\gamma = 0.2$ \\
        $\delta_0 = 1/16 $ 
    \end{tabular}} \\
    Yelp \cite{zeng2019graphsaint} & $716,847$ & $6,977,410$ 
        & $300$ & $100$ (m) & $0.75/0.10/0.15$ (i) 
        & 0.9   & 0.3   &  \\
    Reddit \cite{hamilton2017inductive} & $232,965$ & $114,615,892$ 
        & $602$ & $41$ (s)  & $0.01/0.04/0.96$ (t) 
        & 0.5   & 0.5   &  \\
    Amazon \cite{chiang2019cluster} & $2,400,608$ & $123,718,024$
        & $100$ & $47$ (s)  & $0.70/0.15/0.15$ (i)
        & 0.2   & 0.2   &  \\
    MAG \cite{yang2020scaling} & $27,394,820$ & $366,143,207$
        & $200$ & $100$ (m)  & $0.01/0.01/0.99$ (t)
        & 0.5   & 0.5   &  \\
    Papers100M \cite{Hu2020} & $111,059,956$ & $1,615,685,872$ 
        & $128$ & $172$ (s)  & $0.78/0.08/0.14$ (t) 
        & 0.2   & 0.5   &  \\
    \bottomrule
  \end{tabular}
\end{table*}

\fi

\subsection{Complexity Analysis}
\label{ssec:complexity}
We then develop theoretical analysis on the time and memory complexity of \tname{}. We have the following lemma:
\vspace{-1.ex}
\begin{lemma}
    The time complexity of \pprname{} is $O(\sqrt{\frac{m\Vert \bm{x} \Vert_1}{\beta}})$.
    \label{thm:cplx_push}
\end{lemma}
\vspace{-3.5ex}
\begin{proof}
We analyze the two parts of \refalg{alg:featpush} separately. 
The forward push with early termination runs in $O(\Vert \bm{x} \Vert_1/r_{max})$ according to \cite{Andersen2006}. 
For the random walks in \pprname{}, we employ the complexity derived by \cite{Wang2017} as $O(m \cdot r_{max} / \beta)$.
Hence the overall running time of one query in \refalg{alg:featpush} is bounded by $ O \left( \frac{\Vert \bm{x} \Vert_1}{r_{max}} + r_{max} \cdot \frac{m}{\beta} \right). $By applying Lagrange multipliers, the complexity is minimized by selecting $r_{max} = \sqrt{\frac{\beta \Vert \bm{x} \Vert_1}{m}}$. 
%
\end{proof}
According to \reflem{thm:cplx_push}, the time complexity of computing $\hat{\pi}(\bm{x}, \beta)$ with \refalg{alg:featpush} can be bounded by $O(\sqrt{m\Vert \bm{x} \Vert_1/\beta})$.
To get PPR value with absolute error guarantee of $\lambda$, \refalg{alg:featpush} requires a push coefficient $\beta_s = \frac{\lambda^2/\log(2/\phi)}{2\lambda/3+2}$. Then without \basename{}, the time complexity for computing PPR value for each normalized feature vector is bounded by $O(\sqrt{m/\beta_s})$.

When \basename{} applies, let $\theta_{sum} = \sum_{i=1}^{n_B}\theta_i$ denote the proportion of a feature vector $\bm{x}_f$ computed by the base vectors, and the L1 length of the remaining part $\bm{x}'$ is $1-\theta_{sum}$. In \refalg{alg:featreuse}, we compute the remaining part with push coefficient of $(1-\gamma\theta_{sum})\beta_s$, where $0< \gamma \le 1$. 
Recalling that the L1 length of the feature vector is reduced by $\theta_{sum}$ with \basename{}, we can derive that
the time complexity of computing PPR value of  $\bm{x}$ with \basename{} is $O\left(\sqrt{\frac{m(1-\theta_{sum})}{\beta_s(1 - \gamma\theta_{sum})}}\right )$, which is $\sqrt{\frac{1-\theta_{sum}}{1 - \gamma\theta_{sum}}}$ times smaller than those without \basename{}.

For example, if we compute $\theta_{sum}=1/2$ for a vector $\bm{x}_f$ with the base vectors, and set $\gamma = 1/4$, then the complexity of computing the PPR for $\bm{x}_f$ is $O(\sqrt{4m/7\beta_s})$, which is substantially better than the consumption without \basename{} $O(\sqrt{m/\beta_s})$. The overhead of each base vector is $O(\sqrt{4m/\beta_s})$, which is only twice slower than the original complexity. As we select only a few base vectors, the additional overhead produced by computing base vectors is neglectable comparing with the acceleration gained.

When \basename{} applies, the complexity of computing a feature vector is not worse than the complexity without \basename{}, and are equivalent to the latter only when $\theta_{sum} = 0$ (i.e. the feature vector is completely orthogonal with the base vectors).
Therefore in the worst case, the complexity of \tname{} on feature matrix $\bm{X}$ is equivalent to repeating $F$ queries of \refalg{alg:featpush}. 
By setting $\phi = 1/n$, we can derive the time overhead of \tname{} precomputation. 
For the complexity of memory, the usage of a single-query \pprname{} can be denoted as $O(n)$. Hence the precomputation complexity of \tname{} is given by the following theorem:
\vspace{-0.0ex}
\begin{theorem}
    Time complexity of \tname{} precomputation stage is bounded by $O\left( F\sqrt{m \log n}/\lambda \right)$. Memory complexity is $O\left( nF \right)$. 
\end{theorem}
\vspace{-0.6ex}

\ifx\istech\undefined
    \begin{table*}[!t]
\begin{threeparttable}
  \caption{Average results of \tname{} and baselines on large-scale datasets for transductive and inductive learning. ``Learn'' and ``Infer'' columns are the learning (sum of precomputation and training) and inference time ($\si{s}$), respectively. ``Mem.'' is the peak RAM memory ($\si{GB}$). ``F1'' is the micro F1-score ($\%$) on testing sets. ``$\si{OOM}$'' stands for out of memory error, \blue{``$>12\si{h}$'' means the model requires more than $12\si{h}$ clock time to produce proper results.} The respective models of first and second best performance in ``Learn'', ``Infer'', ``Mem.'', and ``F1'' columns are marked in bold and underlined font. }
  \vspace{-0.6ex}
  \label{tab:rest}
  \centering
  \small
  \setlength{\tabcolsep}{2.5pt}
  \begin{tabular}{@{}c|
    r@{ (\hspace{-1pt}}r@{ +\hspace{-3pt}}r@{)\hspace{3pt}}rr@{\hspace{8pt}}r|
    r@{ (\hspace{-1pt}}r@{ +\hspace{-1pt}}r@{)\hspace{3pt}}rr@{\hspace{6pt}}r|
    r@{ (\hspace{-1pt}}r@{ +\hspace{-1pt}}r@{)\hspace{3pt}}rr@{\hspace{6pt}}r@{}} \toprule
    \multirow{2}{*}{\textbf{Transductive}}
    & \multicolumn{6}{c|}{\textbf{Reddit}} & \multicolumn{6}{c|}{\textbf{MAG}} & \multicolumn{6}{c}{\textbf{Papers100M}} \\
    & \multicolumn{1}{c}{\textbf{\footnotesize Learn}} & \multicolumn{1}{l}{\textbf{\scriptsize ( Pre.}} &
        \multicolumn{1}{c}{\textbf{\hspace{-1pt}\scriptsize + Train)}} & \multicolumn{1}{c}{\textbf{\footnotesize Infer}} & \multicolumn{1}{c}{\textbf{\footnotesize Mem.}} & \multicolumn{1}{c|}{\textbf{\footnotesize F1}}
      & \multicolumn{1}{c}{\textbf{\footnotesize Learn}} & \multicolumn{1}{l}{\textbf{\scriptsize ( Pre.}} &
        \multicolumn{1}{c}{\textbf{\hspace{-3pt}\scriptsize + Train)}} & \multicolumn{1}{c}{\textbf{\footnotesize Infer}} & \multicolumn{1}{c}{\textbf{\footnotesize Mem.}} & \multicolumn{1}{c|}{\textbf{\footnotesize F1}}
      & \multicolumn{1}{c}{\textbf{\footnotesize Learn}} & \multicolumn{1}{l}{\textbf{\scriptsize ( Pre.}} &
        \multicolumn{1}{c}{\textbf{\hspace{-1pt}\scriptsize + Train)}} & \multicolumn{1}{c}{\textbf{\footnotesize Infer}} & \multicolumn{1}{c}{\textbf{\footnotesize Mem.}} & \multicolumn{1}{c}{\textbf{\footnotesize F1}} \\
    \midrule
    GraphSAINT &
        \ul{51.5} & \multicolumn{1}{c}{--} &  51.5 & 26.1 & 11.1 & 30.7 \textpm 3.0 &
      \multicolumn{1}{c}{--} & \multicolumn{1}{c}{--} & \multicolumn{1}{c}{--} & \multicolumn{1}{c}{--} & \multicolumn{1}{c}{\scriptsize OOM\,} & \multicolumn{1}{c|}{--} &
      \multicolumn{1}{c}{--} & \multicolumn{1}{c}{--} & \multicolumn{1}{c}{--} & \multicolumn{1}{c}{--} & \multicolumn{1}{c}{\scriptsize OOM\,} & \multicolumn{1}{c}{--} \\
    GAS &
        3563 & \multicolumn{1}{c}{--} &  3563 &   \textbf{0.1} &  14.6 & 38.0 \textpm 0.2 &
      \multicolumn{1}{c}{--} & \multicolumn{1}{c}{--} & \multicolumn{1}{c}{--} & \multicolumn{1}{c}{--} & \multicolumn{1}{c}{\scriptsize OOM\,} & \multicolumn{1}{c|}{--} &
      \multicolumn{1}{c}{--} & \multicolumn{1}{c}{--} & \multicolumn{1}{c}{--} & \multicolumn{1}{c}{--} & \multicolumn{1}{c}{\scriptsize OOM\,} & \multicolumn{1}{c}{--} \\
    PPRGo &
         163 &   157 &   4.8 &  74.1 &   \ul{8.0} & 31.0 \textpm 1.7 &
       \multicolumn{1}{c}{--} & \multicolumn{1}{c}{$>12\si{h}$} & \multicolumn{1}{c}{--} & \multicolumn{1}{c}{--} & \ul{146} & \multicolumn{1}{c|}{--} &
       \multicolumn{1}{c}{--} & \multicolumn{1}{c}{--} & \multicolumn{1}{c}{--} & \multicolumn{1}{c}{--} & \multicolumn{1}{c}{\scriptsize OOM\,} & \multicolumn{1}{c}{--} \\
    GBP &
        1891 &  2127 &  16.3 &   6.2 &  8.4 & \ul{39.2 \textpm 0.3} &
        \ul{4572} &  4470 &   102 &  \ul{1433} &  177\,\tnote{$\ast$} & \ul{34.8 \textpm 0.1} &
      \multicolumn{1}{c}{--} & \multicolumn{1}{c}{--} & \multicolumn{1}{c}{--} & \multicolumn{1}{c}{--} & \multicolumn{1}{c}{\scriptsize OOM\,} & \multicolumn{1}{c}{--} \\
    \textbf{\tname{} (ours)} &
      \textbf{  12.0} &   1.8 &  10.6 & \ul{ 4.8} & \textbf{ 4.7} & \textbf{40.3 \textpm 0.7} &
      \textbf{   460} &   380 &  80.0 & \textbf{1421} & \textbf{49.4} & \textbf{35.0 \textpm 0.3} &
      \textbf{  1471} &  83.5 &  1388 & \textbf{ 2.8} & \textbf{63.7} & \textbf{35.5 \textpm 0.8} \\
    \midrule
    \multirow{2}{*}{\textbf{Inductive}} &
    \multicolumn{6}{c|}{\textbf{PPI}} & \multicolumn{6}{c|}{\textbf{Yelp}} & \multicolumn{6}{c}{\textbf{Amazon}} \\
    & \multicolumn{1}{c}{\textbf{\footnotesize Learn}} & \multicolumn{1}{l}{\textbf{\scriptsize ( Pre.}} &
        \multicolumn{1}{c}{\textbf{\hspace{-1pt}\scriptsize + Train)}} & \multicolumn{1}{c}{\textbf{\footnotesize Infer}} & \multicolumn{1}{c}{\textbf{\footnotesize Mem.}} & \multicolumn{1}{c|}{\textbf{\footnotesize F1}}
      & \multicolumn{1}{c}{\textbf{\footnotesize Learn}} & \multicolumn{1}{l}{\textbf{\scriptsize ( Pre.}} &
        \multicolumn{1}{c}{\textbf{\hspace{-3pt}\scriptsize + Train)}} & \multicolumn{1}{c}{\textbf{\footnotesize Infer}} & \multicolumn{1}{c}{\textbf{\footnotesize Mem.}} & \multicolumn{1}{c|}{\textbf{\footnotesize F1}}
      & \multicolumn{1}{c}{\textbf{\footnotesize Learn}} & \multicolumn{1}{l}{\textbf{\scriptsize ( Pre.}} &
        \multicolumn{1}{c}{\textbf{\hspace{-1pt}\scriptsize + Train)}} & \multicolumn{1}{c}{\textbf{\footnotesize Infer}} & \multicolumn{1}{c}{\textbf{\footnotesize Mem.}} & \multicolumn{1}{c}{\textbf{\footnotesize F1}} \\
    \midrule
    GraphSAINT &
        2813 & \multicolumn{1}{c}{--} &  2813 &  4.1 &  3.2 & 89.3 \textpm 0.2 &
        8589 & \multicolumn{1}{c}{--} &  8589 &  193 & 54.0 & \textbf{64.9 \textpm 0.1} &
        2612 & \multicolumn{1}{c}{--} &  2612 &  804 & 87.9 & 81.3 \textpm 0.1 \\
    GAS &
         326 & \multicolumn{1}{c}{--} &   326 & \textbf{ 0.1} &  6.6 & \ul{99.3 \textpm 0.1} &
        3622 & \multicolumn{1}{c}{--} &  3622 & \textbf{ 0.1} & 22.0 & \ul{63.8 \textpm 0.0} &
       19218 & \multicolumn{1}{c}{--} & 19218 & \textbf{ 0.4} & 41.7 & 71.7 \textpm 0.5 \\
    PPRGo &
        4019 &  70.0 &   3949 &   1.7 &   2.7 & 50.1 \textpm 0.7 &
       13073 &  91.9 &  12981 &  30.1 &  16.9 & 56.5 \textpm 2.6 &
        3041 &  2092 &    949 &  63.3 &  27.4 & 78.4 \textpm 3.0 \\
    GBP &
        \ul{86.4} &  18.5 &  67.9 &  0.3 & \textbf{ 2.5} & 99.3 \textpm 0.0 &
         \ul{198} &  77.2 &   121 & \ul{ 2.9} & \ul{13.4} & 60.6 \textpm 0.1 &
        \ul{2193} &  1019 &  1174 &  7.5 & \ul{13.4} & \textbf{86.8 \textpm 0.1} \\
    \textbf{\tname{} (ours)} &
      \textbf{  49.3} &   0.5 &  48.9 & \ul{ 0.3} &  \ul{2.5} & \textbf{99.3 \textpm 0.0} &
      \textbf{   154} &   3.6 &   150 &  3.1 & \textbf{ 7.4} & 61.4 \textpm 0.4 &
      \textbf{  1281} &   7.0 &  1274 & \ul{6.8} & \textbf{ 7.3} & \ul{83.8 \textpm 0.1} \\
    \bottomrule
  \end{tabular}
  \begin{tablenotes}
    \footnotesize
    \item [$\ast$] GBP experiment of this entry is conducted on a different machine with a larger 192GB RAM. 
  \end{tablenotes}
\end{threeparttable}
\end{table*}

\else

\fi


\section{Experimental Evaluation}
\label{sec:experiment}

\blue{In this section we compare the \tname{} performance with other scalable GNN competitors under similar parameter settings.  }

\subsection{Experiment Setting}
\label{ssec:setting}

\noindentparagraph{Datasets. }
We adopt benchmark datasets of different graph properties, feature dimensions, and data splitting for large-scale node classification tasks. We present the dataset statistics in \reftab{tab:dataset}. 
Among the datasets, PPI, Yelp, and Amazon are for \textit{inductive} learning, where the training and testing graphs are different and require separate graph precomputation and propagation. The given original node splittings are in \reftab{tab:dataset}. 
The learning tasks on the other datasets are \textit{transductive} and are performed on the same graph structure. For a dataset with $N_c$ target classes, we refer to convention in \cite{kipf2016semi,Bojchevski2020} to randomly select two sets of $20 N_c$ and $200 N_c$ nodes for training and validation, respectively, and the rest labeled nodes in the graph as the testing set. 

\noindentparagraph{Metrics. }
Predictions on datasets PPI, Yelp, and MAG are multi-label classification having multiple targets for each node. The other tasks are multi-class with only one target class per node. We uniformly utilize micro F1-score to assess the model prediction performance. 
The evaluation is conducted on a machine with Ubuntu 18 operating system, with 160GB RAM, an Intel Xeon CPU (2.1GHz), and an NVIDIA Tesla K80 GPU (11GB memory). The implementation is by PyTorch and C++. 

\noindentparagraph{Baseline Models. }
We select the state-of-the-art models of different scalable GNN methods analyzed in \refsec{sec:preliminary} as our baselines. 
GraphSAINT-RW \cite{zeng2019graphsaint} and GAS \cite{Fey2021} are representative of different sampling-based algorithms. For post- and pre-propagation decoupling approaches, we respectively employ the most advanced PPRGo \cite{Bojchevski2020} and GBP \cite{Chen2020a}. 
For a fair comparison, we mostly retain the implementations and settings from original papers and source codes.
We uniformly apply single-thread executions for all models. 

\ifx\istech\undefined
    \begin{figure}[bh]
    \centering
    \subcaptionbox{Precomputation Time\label{fig:param1}}%
    [0.46\linewidth]{\includegraphics[height=1.3in]{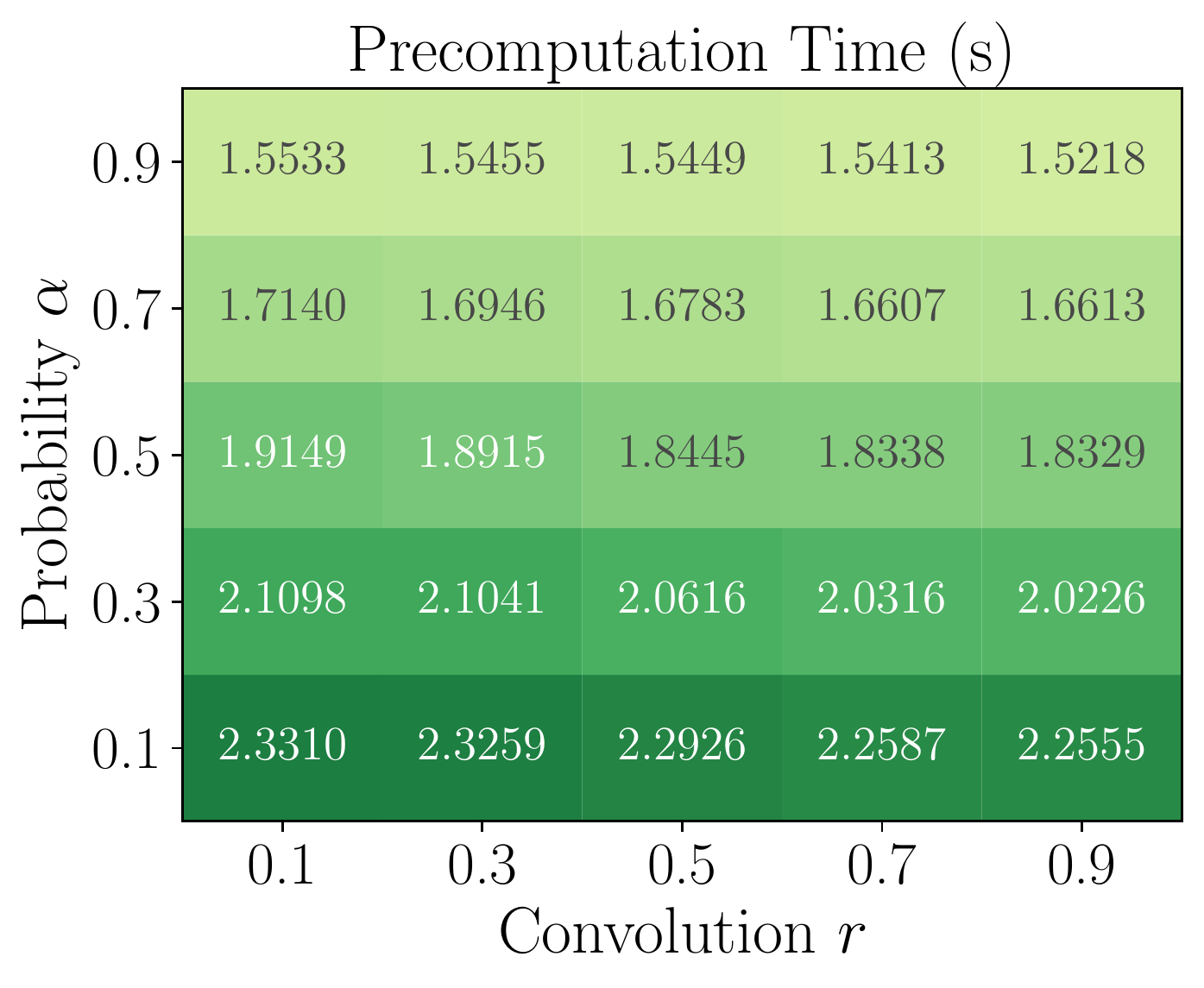}}
    \hfil
    \subcaptionbox{Testing Accuracy\label{fig:param2}}%
    [0.46\linewidth]{\includegraphics[height=1.3in]{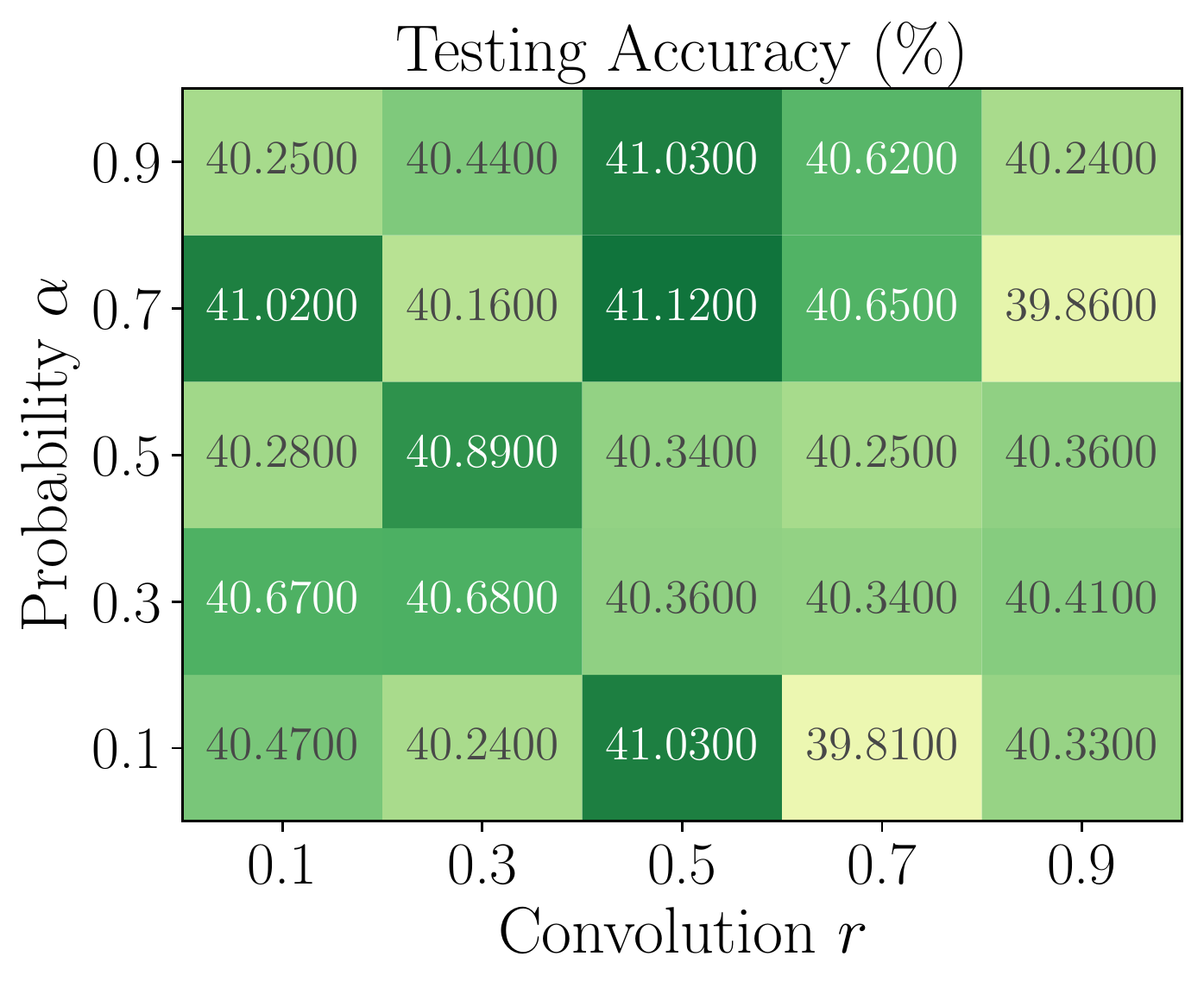}}
    \vspace{-1ex}
    \caption{\blue{Effect of propagation parameters $\alpha$ and $r$ on \tname{} \textbf{(a)} efficiency and \textbf{(b)} accuracy for Reddit dataset. }}
  \label{fig:param}
\end{figure}

\fi

\noindentparagraph{Hyperparameters. }
Propagation parameters $\alpha, r, \lambda$ and \rename{} parameters $n_B, \gamma, \delta_0$ are presented in \reftab{tab:dataset} per dataset. 
For neural network architecture we set layer depth $L=4$, layer width $W=2048$ and $W=128$ for inductive and transductive tasks, respectively, to be aligned with optimal baseline results in \cite{Chen2020a}. 
In model optimization, we employ mini-batch training with respective batch size $2048$ and $64$ for inductive and transductive learning, for a maximum of $1000$ epochs with early stopping. 

\ifx\istech\undefined
    \blue{The effects of $\alpha$ and $r$ values on particular metrics are presented in \reffig{fig:param}, which demonstrates that our selection of parameter values are efficient and do not influence GNN performance. The full exploration on hyperparameters can be found in \cite{techreport}. }
\fi

\subsection{Performance Comparison}
\label{ssec:performance}
\begin{figure*}[!t]
    \centering
    \subcaptionbox{Reddit\label{fig:conv_reddit}}%
    [0.30\linewidth]{\includegraphics[height=1.18in]{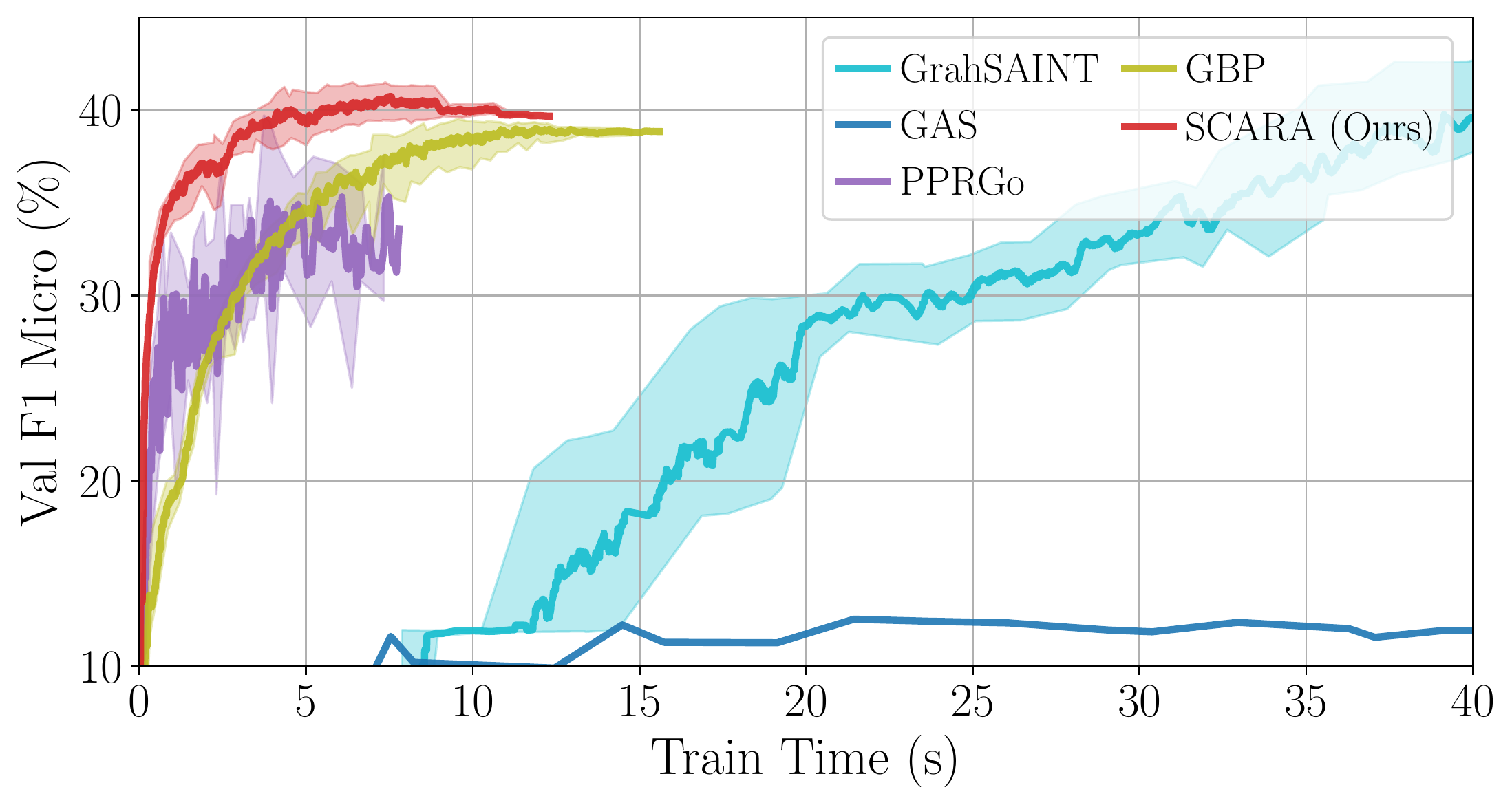}}
    \hfil
    \subcaptionbox{PPI\label{fig:conv_ppi}}%
    [0.30\linewidth]{\includegraphics[height=1.18in]{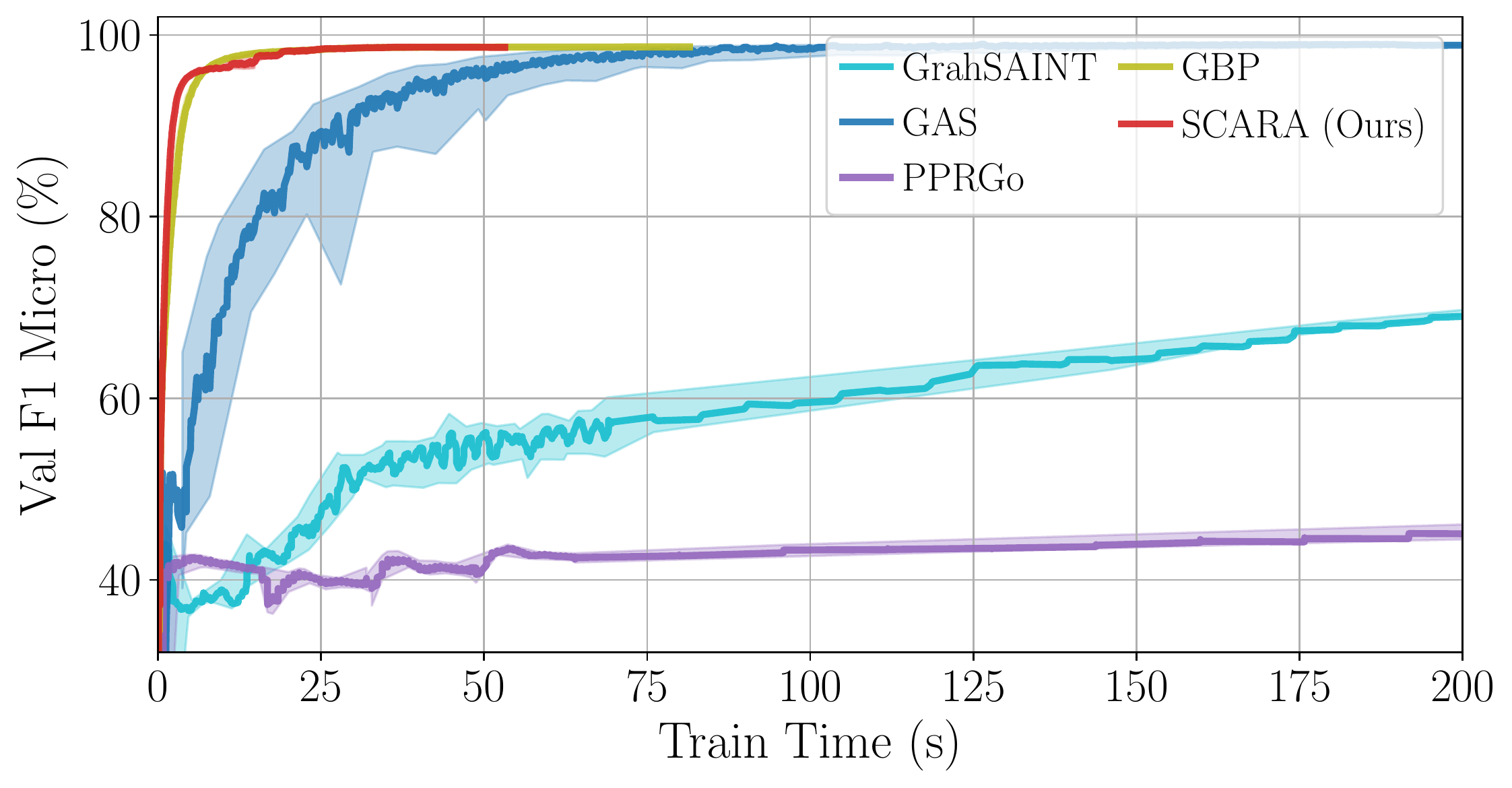}}
    \hfil
    \subcaptionbox{Yelp\label{fig:conv_yelp}}%
    [0.30\linewidth]{\includegraphics[height=1.18in]{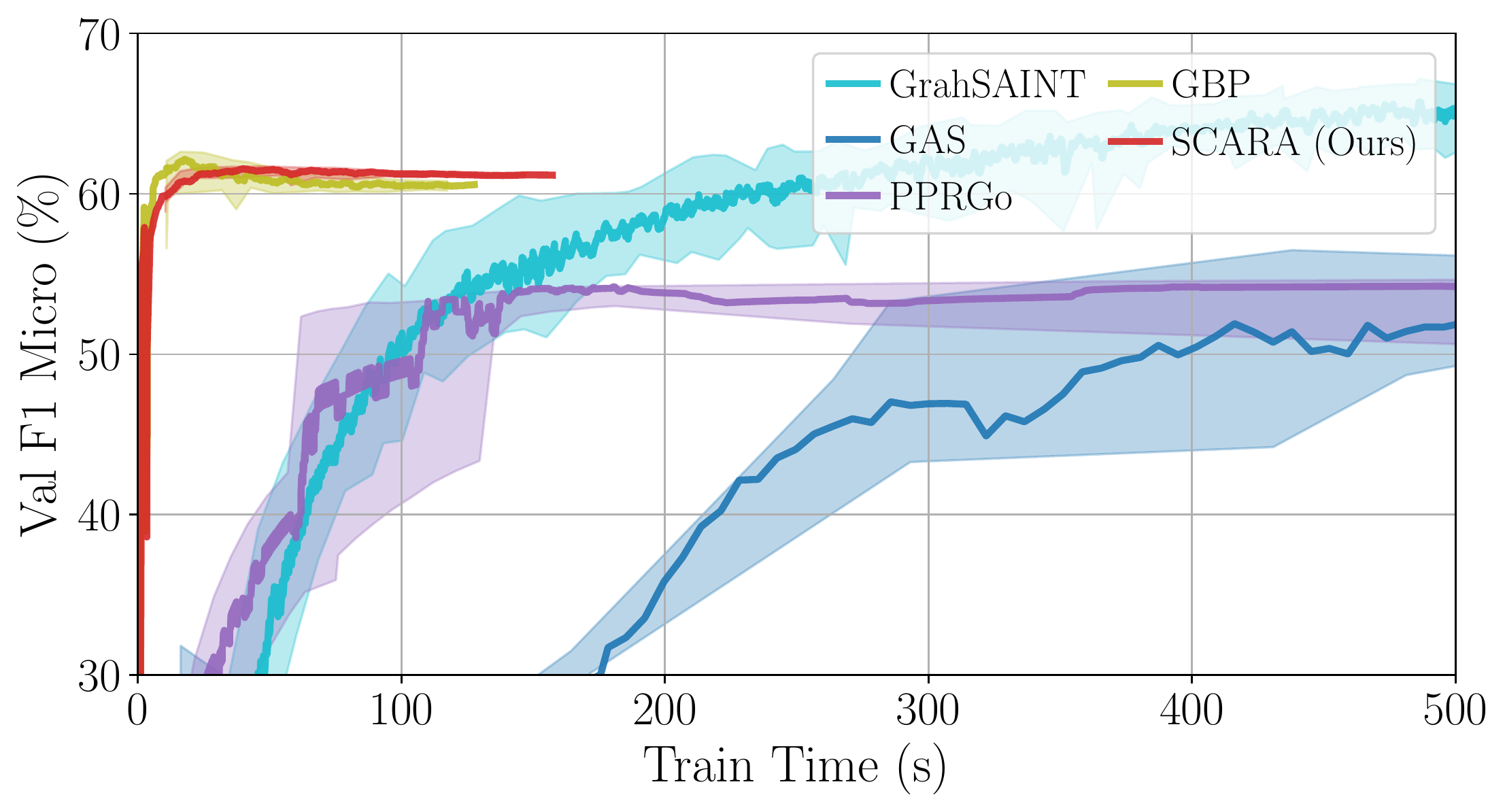}}
    \vspace{-1ex}
    \caption{\blue{Validation F1 convergence curves of \tname{} and baseline models on \textbf{(a)} Reddit, \textbf{(b)} PPI, and \textbf{(c)} Yelp datasets. Curves only represents the process of training phase. Shaded area is result range of multiple runs. }}
  \label{fig:conv}
\vspace{-2.ex}
\end{figure*}

We evaluate the performance of \tname{} and baselines in terms of both effectiveness and efficiency. 
\blue{\reftab{tab:rest} shows the average results of repetitive experiments on 6 large datasets, including the assessments on accuracy, memory, and the running time for different phases. Among them the key metric is learning time, which is summed up by precomputation and training times and presents the efficiency through the information retrieving process to acquire an effective model. }
The training curves is  given in \reffig{fig:conv}. 

As an overview, the experimental results demonstrate the superiority of our model achieving scalability through the learning phases. On all benchmark datasets, \tname{} reaches $5 - 100\times$ acceleration in precomputation time, as well as comparable or better training and inference speed, and significantly better memory overhead. When the graphs are scaled-up, the time and memory footprints of \tname{} increase relatively slower than our GNN baselines, which is in line with our complexity analysis. 
For prediction performance, \tname{} converges in all tasks and outputs comparable or better accuracy than other scalable competitors. 

From the aspect of time efficiency, our \tname{} model effectively speeds up the learning process in all tasks, mostly benefited by the fast and stable precomputation for graph propagation. The simple neural model forwarding implemented in mini-batch approach also contributes to the efficient computation of model training and inference. 
On the largest available dataset Papers100M, our method efficiently completes precomputation in 100 seconds, and finishes learning in an acceptable length of time, showing the scalability of processing billion-scale graphs. 
\blue{In comparison, sampling-based GraphSAINT and GAS achieves good performance on several datasets, but the $O(ILmF)$ term in training complexity results in great slowdown when the graph is scaled-up. GraphSAINT is costly for its full-batch prediction stage on the whole graph, which is usually only executable on CPU. GAS is particularly fast for inference, but which comes with the price of trading off memory expense and training time to manipulate these cache. 
The propagation decoupling models PPRGo and GBP show better scalability, but take more time than \tname{} to converge, due to the graph information yielded by precomputation algorithms. It can be seen that their node-based propagation computations are less efficient when the graph sizes become larger, which aligns with \reftab{tab:complexity} complexity analysis.}
Remarkably, \tname{} achieves about $100\times$ and $40\times$ faster for precomputation than these two competitors on Reddit and PPI. 

When considering memory overhead, our method also demonstrates its efficiency benefit from our scalable implementation. We discover that the major memory expense of \tname{} only increases proportional to the graph attribute matrix, while PPRGo and GBP usually demand twice as large RAM, and GraphSAINT and GAS use even more for their samplers. On the billion-scale Papers100M graph, most baselines meet out of memory error in our machine.

For learning effectiveness, \tname{} achieves similar or better F1-score compared with current GNN baselines. For most datasets, our model outperforms both the state-of-the-art pre-propagation approach GBP and the scalable post-propagation baseline PPRGo. 
It is worth noting that most baselines fail to or only partially converge before training terminates in certain tasks. 

\reffig{fig:conv} shows the validation F1-score versus training time on representative datasets and corresponding GNN models. It can be observed that when comparing the time consumption to convergence, the \tname{} model is efficient in reaching the same precision faster than most methods. 
\blue{The performances of several baselines in the figure are suboptimal because they require more training time beyond the display scopes to converge. }

\ifx\istech\undefined
\else
    \subsection{Baseline Comparison}
In this section, we further explain our \tname{} method comparing the baseline scalable GNNs. 
For inference stage, by design the speed between our \tname{} model and GBP should be identical, as both utilize the same strategy of simply performing neural network feature transformation without propagation. Actually it is the optimal inference efficiency in the context of GNN, benefit from the decoupled beforehand propagation, and is favorable for applications such as online settings because it no longer requires interacting with graph structure after deployment. 
\reftab{tab:rest} shows that in all applicable datasets, the recorded inference time is at the same level of GBP, and the difference is within the error margin of repetitive experiments. It also indicates that the inference efficiency of both \tname{} and GBP are significantly better than other non-pre-propagation approaches, hence ideal for practical use. 

\begin{table}[bp]
  \caption{Comparison between \tname{} and GBP. }
  \vspace{-0.8ex}
  \label{tabr:compgbp}
  \centering
  \small
  \setlength{\tabcolsep}{3.5pt}
  \begin{tabular}{@{}c|cccc@{}} \toprule
    \multirow{2}{*}{\textbf{Model}} &  \multirow{2}{*}{\textbf{Propagation}} 
        & \textbf{Sensitive} & \textbf{Computation} & \textbf{Memory} \\
     &  & \textbf{to $n$}    & \textbf{Reuse}       & \textbf{Overhead} \\
    \midrule
    \textbf{GBP}      & Node-oriented        & Yes  & N/A        & $O(4nF)$ \\
    \textbf{\tname{}} & Feature-oriented     & Less & \rename{}  & $O(2nF)$ \\
    \bottomrule
  \end{tabular}
\end{table}

Apart from inference queries, the learning stage is also important in scalable GNN research. The majority of related studies \cite{chiang2019cluster,zeng2019graphsaint,Klicpera2019} focus on improving the learning time, as it is dominant in the resource-demanding GNN learning pipeline. 
Typical GNN training performs iterative updates that query every node in graph for $I$ epochs, where $I = 1000$ in our experiments. In such case, effectively acquiring a learned model is prohibited if the learning expense is too high. Our pre-propagation solution moves propagation outside iterative training, hence the latter becomes simple and efficient. 

In particular, with regard to GBP, which is the state-of-the-art in pre-propagation GNNs with enhanced propagation and precision over SGC \cite{wu19sim}, we make a take-away table for our novelty in \reftab{tabr:compgbp}. Comparing to GBP, our \tname{} model greatly optimizes the pre-computation procedure as highlighted in \reftab{tabr:compgbp}, hence produces better overall efficiency. 

The GBP model employs a PPR-based bidirectional propagation from both node and feature dimensions. It hence still contains a node-oriented scheme. Interpret from complexity analysis in \refsec{sec:preliminary}, it is sensitive to the graph size $n$ and slows down significantly for large-scale graphs. In comparison, our Forward Push and Random Walk both operate in the usually constant feature dimension and produces better efficiency and scalability. To study the approximation on features and utilize generalized PPR calculations, we uniquely propose a complete theoretical framework for precision and complexity analysis for feature PPR problem in \refsec{ssec:fwdpush}. This distinguish \tname{} from GBP which relies on straightforward PPR algorithms. Benefit from the pure feature-oriented calculation, \tname{} is able to perform optimizations based on particular properties of feature vectors, illustrated as \rename{} in our \refsec{ssec:freuse}. Such technique is not applicable on GBP propagation due to its interdependence of nodes among features. 

\begin{table}[tp]
  \caption{Comparison of \tname{} and GBP precomputation on MAG and Papers100M on alternative machine. }
  \vspace{-0.8ex}
  \label{tabr:paper}
  \centering
  \begin{tabular}{@{}c|
    rrrr@{}} \toprule
    \multirow{2}{*}{\textbf{Transductive}}
    & \multicolumn{2}{c}{\textbf{MAG}} & \multicolumn{2}{c}{\textbf{Papers100M}} \\
    & \multicolumn{1}{c}{\textbf{\footnotesize Pre.}} & \multicolumn{1}{c}{\textbf{\footnotesize Mem.}} & \multicolumn{1}{c}{\textbf{\footnotesize Pre.}} & \multicolumn{1}{c}{\textbf{\footnotesize Mem.}}  \\
    \midrule
    GBP &
      $4470\si{s}$ & $ 177\si{GB}$ & \multicolumn{1}{c}{--} & \multicolumn{1}{c}{\small OOM\,} \\
    \textbf{\tname{} (ours)} &
      $ 449\si{s}$ & $49.4\si{GB}$ & $73.6\si{s}$ & $63.7\si{GB}$ \\
    \bottomrule
  \end{tabular}
\end{table}

We conduct additional experiments on another machine which has larger RAM capacity ($256\si{GB}$) to particularly compare the precomputation overhead between \tname{} and GBP on the largest MAG and Papers100M. 
\reftab{tabr:paper} shows that on MAG, GBP requires more than $10\times$ processing time as well as 177GB RAM, which exceeds our original machine with 160GB bound. For Papers100M which has about $40\si{GB}$ raw data, even the $256\si{GB}$ capacity is insufficient for GBP to complete precomputation. In fact, we note that in \cite{Wang2021}, it utilizes up to 512GB RAM to conduct the experiment of GBP on Papers100M. 

We compare the estimated memory overhead between \tname{} and GBP in \reftab{tabr:compgbp}. While both models have an $O(nF)$ complexity, the hidden coefficients are different. We conclude two major causes by analyzing the source code of GBP: Firstly, GBP adopts node-oriented processing on the attribute matrix, which naturally requires more intermediate cache; Secondly, GBP practically employs higher numeric precision in storing and calculating the features, which is proved by us does not impact the accuracy in approximate propagation in the following section. Note that the numeric precision is different from (and much smaller than) the algorithmic precision $\lambda$, with the latter we control the same among models. 
Such analysis explains the performance of GBP on large datasets as demonstrated in \reftab{tabr:paper}. 
    \subsection{Effect of Parameters}
In this section we explain the selection of parameters. 
Regarding the default selection in \reftab{tab:dataset}, we set the values of $\alpha$ and $r$ for shared datasets mainly according to GBP \cite{Chen2020a,Wang2021} in order to produce comparable results. Reddit and MAG are employed with our own strategy: we use $r = 0.5$ for better comparison and generality, while $\alpha$ is decided based on graph density.  
The error bound $\lambda$ can be arbitrarily large as long as it does not reduce effectiveness, we hence uniformly set it to $\lambda = 1 \times 10^{-4}$ for all datasets to provide aligned evaluations across datasets. 

We conduct a grid search on the value ranges of $\alpha$ and $r$ as shown in \reffig{figr:param_push}. It can be inferred that a larger $\alpha$ has a slight improvement on precomputation efficiency, while $r$ has no significant impact. The accuracy is relatively not sensitive with variance inside the error range ($\pm 1\%$), proving that the model is robust to the changes of both $\alpha$ and $r$. We hence conclude that the parameters can be determined without requiring a sophisticated tuning. 

\begin{figure}[!b]
    \centering
    \subcaptionbox{Precomputation Time\label{fig:param_push1}}%
    [0.46\linewidth]{\includegraphics[height=1.6in]{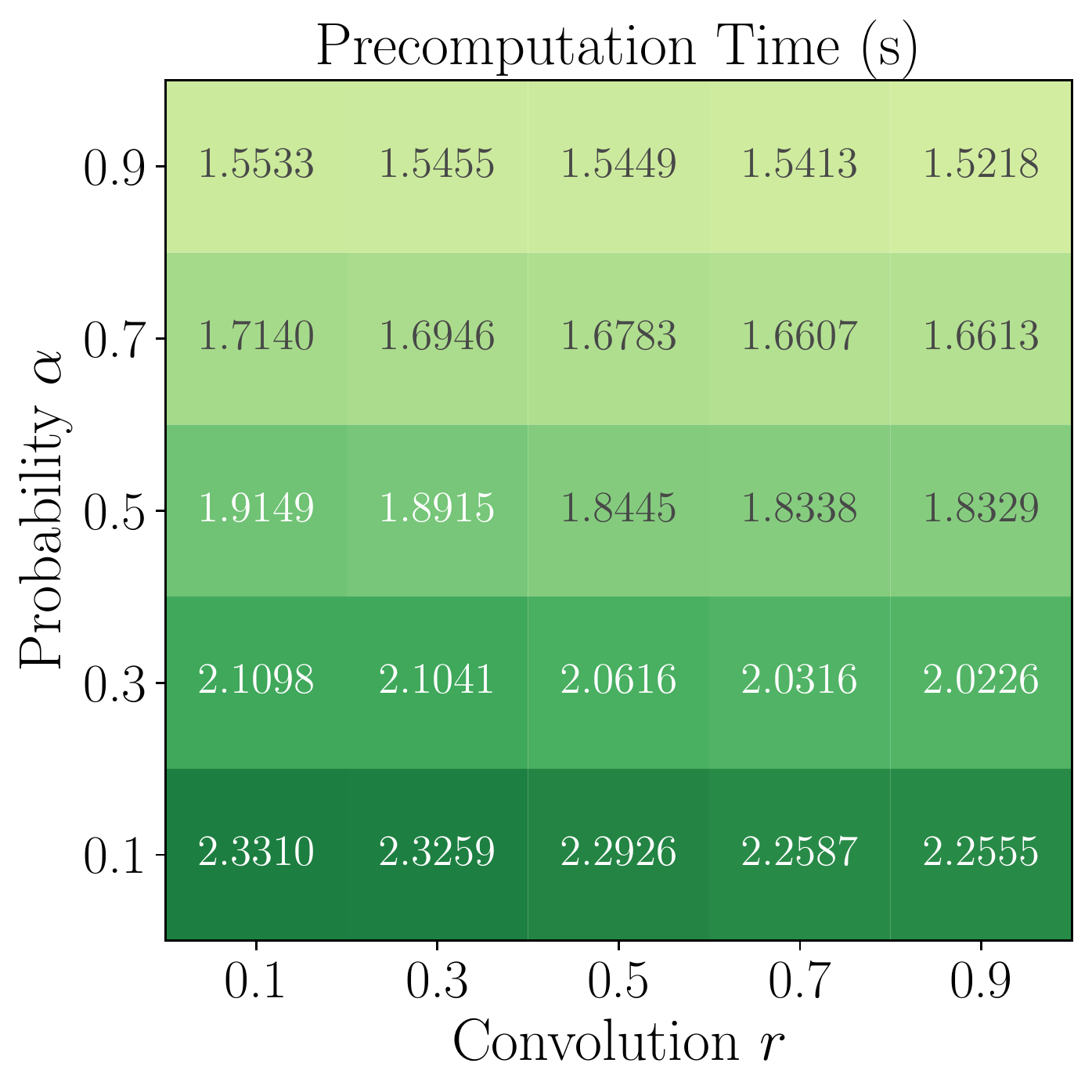}}
    \hfil
    \subcaptionbox{Testing Accuracy\label{fig:param_push2}}%
    [0.46\linewidth]{\includegraphics[height=1.6in]{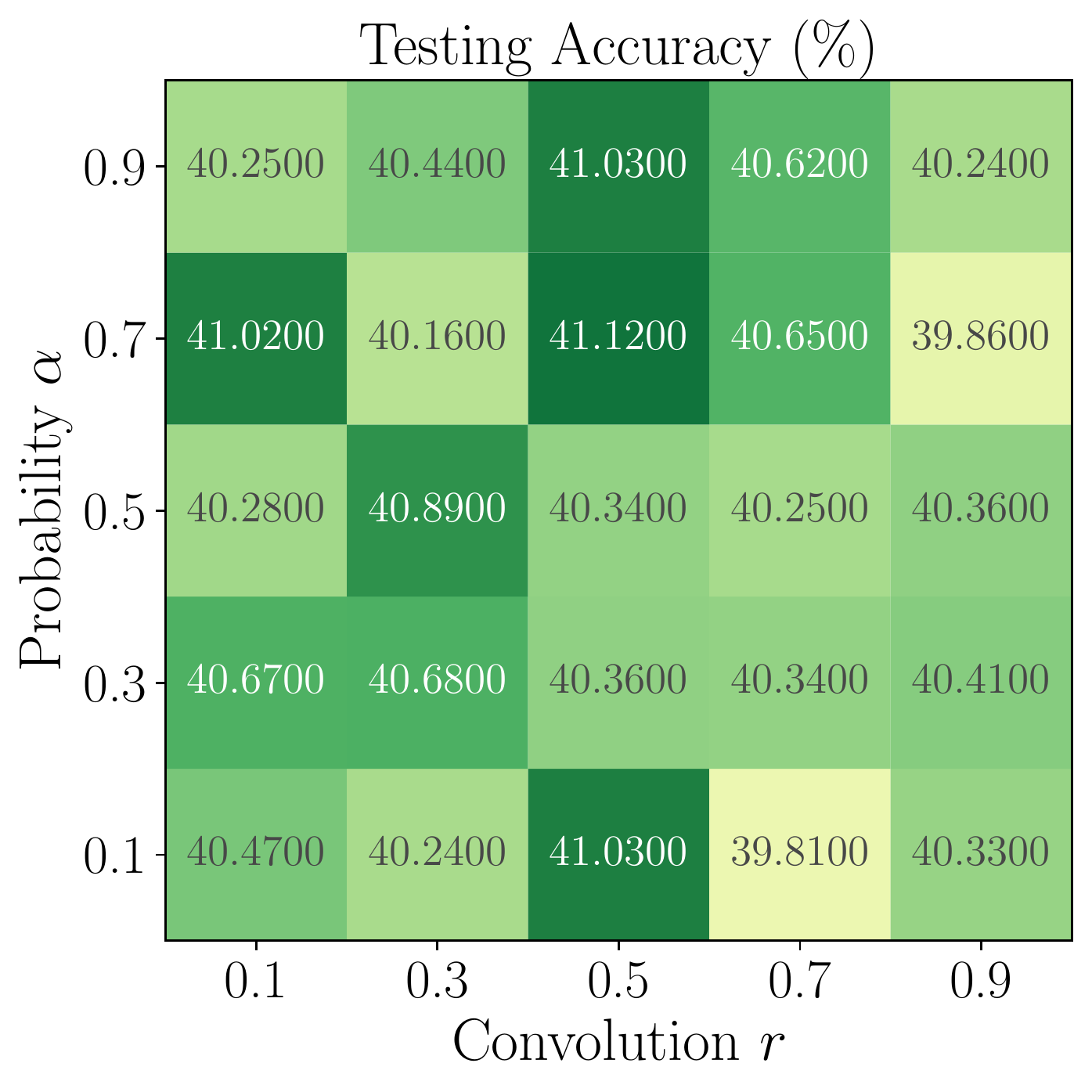}}
    \vspace{-1ex}
    \caption{Effect of teleport probability $\alpha$ and convolution coefficient $r$ values on \tname{} \textbf{(a)} precomputation time and \textbf{(b)} testing accuracy for Reddit dataset. }
  \label{figr:param_push}
\end{figure}

Here we would like to explain the reason behind our propagation parameter selection, especially in \reftab{tab:rest}, in two folds. Theoretically, the values of the three parameters affect the propagation from different aspects. 
Intuitively, $\alpha$ is the PPR teleport probability of \pprname{}, which is dependent on graph adjacency. $\alpha$ is usually set larger to mitigate density for graphs with higher average degrees \cite{Bojchevski2020}. 
The factor $r$ controls the normalization strength of the degree matrix $\bm{D}$ as shown in \refeq{eq:overview_p}. Especially, when $r = 0.5$, it degrades to the normalized adjacency matrix $\Tilde{\bm{A}}$ presented in APPNP\cite{Klicpera2019} and PPRGo\cite{Bojchevski2020}. 
The error bound $\lambda$ determines the approximation push coefficient $\beta$ in \refalg{alg:featpush}. Hence, $\lambda$ is used to configure the trade-off between precision and speed in precomputation and tends to be larger for better efficiency. 

Empirically, the values of similar parameters vary greatly even for a same setting in previous works \cite{Klicpera2019,wu19sim,Bojchevski2020,Chen2020a}. While \reffig{figr:param_push} may indicate the eventual performance is not sensitive to parameter selection, we notice the importance of providing comparable results to other works by using the same parameters. We mainly refer to GBP \cite{Chen2020a} on the values of $\alpha$ and $r$ as both belong to the pre-propagation category. 

We conduct additional experiments exploring the \rename{} sensitivity of base size $n_B$ and precision factor $\gamma$. As stated that the neural network is relatively robust and patterns are hard to infer from the testing accuracy, we thence purely investigate the propagation stage. We use the embedding difference, which is calculated by the average absolute difference of each element in the embedding matrix $\bm{P}$ comparing with \tname{} without \rename{}, as the indicator of precision. 

\reffig{figr:param_re} presents the result on precomputation time and precision on Reddit dataset. For comparison, precomputation without \rename{} uses $2.84 \si{s}$. It can be observed that both $n_B$ and $\gamma$ influence \rename{} efficiency for less than $\pm 0.2 \si{s}$. The difference of embedding values is at the level of $10^{-7}$, which is significantly smaller than the algorithmic error bound $\lambda = 10^{-4}$. Experiments on other datasets show similar results. 
We hence conclude that our parameter values of $n_B/n_U = 0.02$ and $\gamma = 0.2$ are effective. 

\begin{figure}[!b]
    \centering
    \subcaptionbox{Precomputation Time\label{fig:param_re1}}%
    [0.46\linewidth]{\includegraphics[height=1.6in]{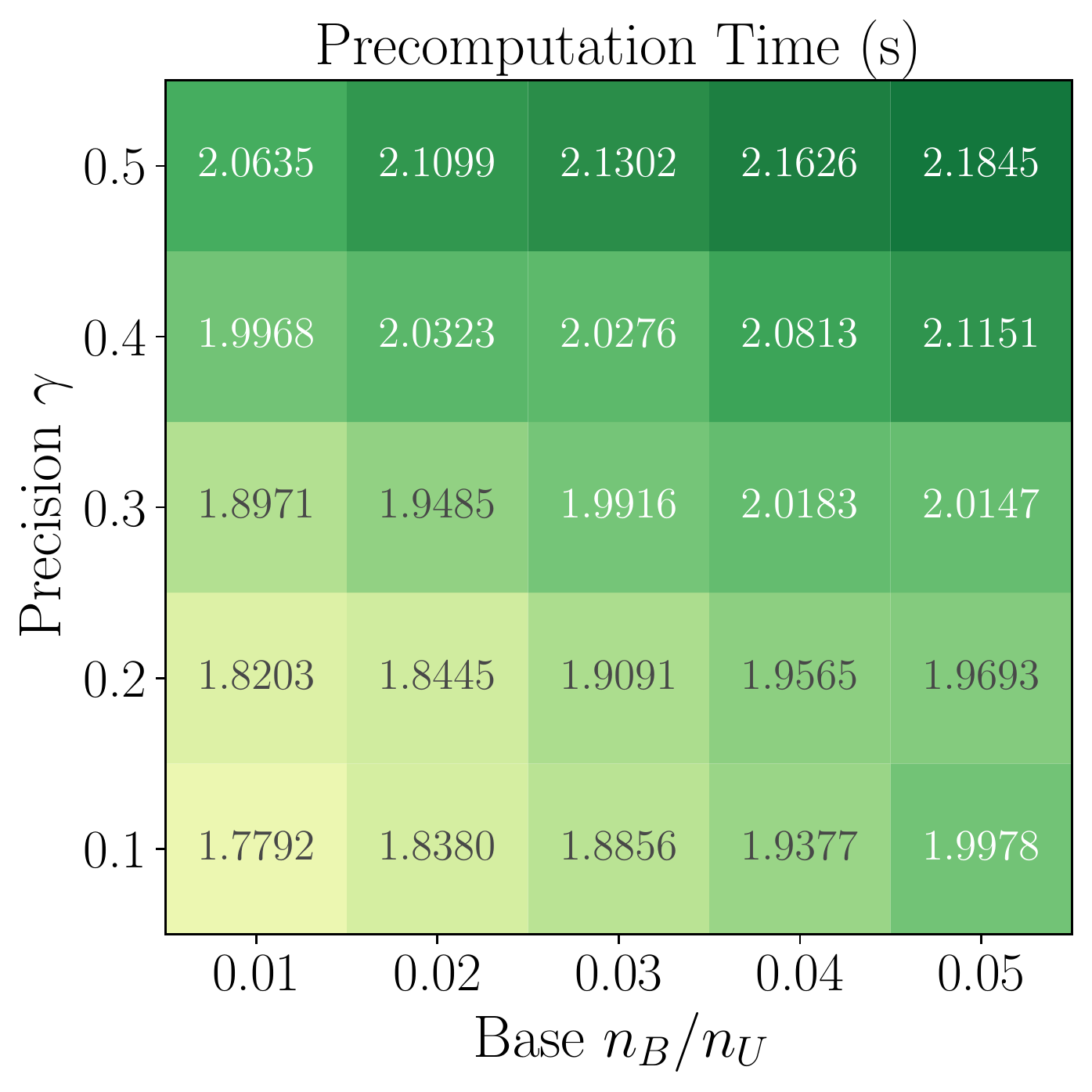}}
    \hfil
    \subcaptionbox{Embedding Difference\label{fig:param_re2}}%
    [0.46\linewidth]{\includegraphics[height=1.6in]{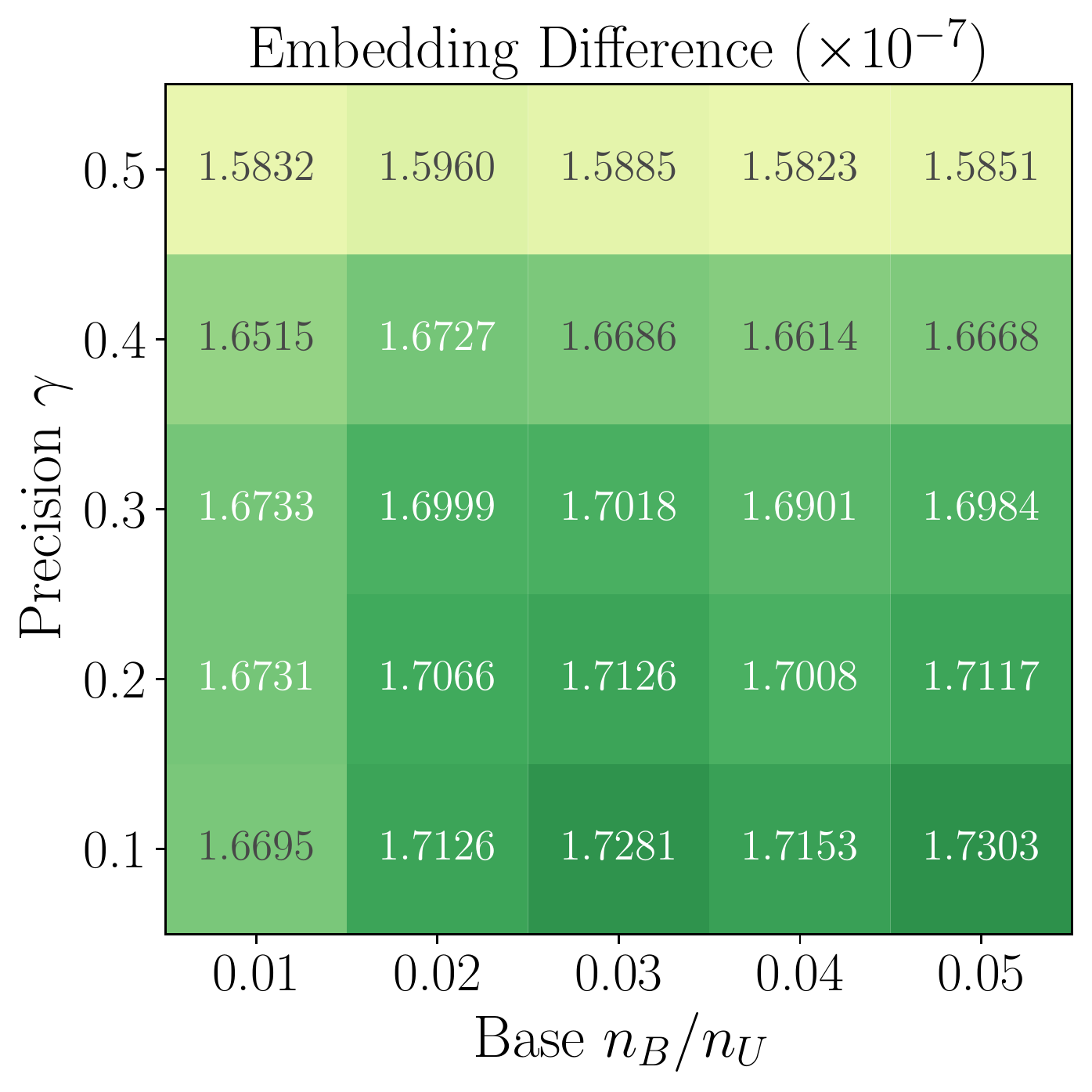}}
    \vspace{-1ex}
    \caption{Effect of precision factor $\gamma$ and base set size $n_B$ values on \tname{} \textbf{(a)} precomputation time and \textbf{(b)} average embedding value difference for Reddit dataset. }
  \label{figr:param_re}
\end{figure}

\begin{table}[tp]
  \caption{Effect of \tname{} decomposition threshold $\delta_0$ on average embedding value difference ($\times 10^{-7}$) for Reddit dataset. }
  \vspace{-0.8ex}
  \label{tabr:delta}
  \centering
  \begin{tabular}{@{}c|rrrrr@{}} \toprule
    \textbf{Threshold $\bm{\delta_0}$} & 
        $\bm{1}$ & $\bm{1/2}$ & $\bm{1/4}$ & 
        $\bm{1/16}$ & $\bm{1/64}$ \\
    \midrule
    \textbf{Embedding Difference} &
        20.40 & 0.46 & 0.45 & 0.45 & 0.45 \\
    \bottomrule
  \end{tabular}
\end{table}

We then explore the effect of decomposition threshold $\delta_0$ by experiments. Results of the embedding value difference are shown in \reftab{tabr:delta}. Even with $\delta_0 = 1$, which the decomposition loop in \refalg{alg:featreuse} executes only once for each feature, the decomposition error on embedding value is still less than $\lambda = 10^{-4}$. When $\delta_0 < 1/2$, the error gradually converges, and we safely set threshold $\delta_0 = 1/16$. 
Regarding efficiency, the loop uses less than $10^{-4} \si{s}$ per feature, hence has no significant influence on speed. 

The current network architecture for \reftab{tab:rest} empirical evaluation is the same for transductive and inductive learning tasks, as described in \refsec{ssec:setting}. 
In this paper we mainly study and enhance the process of propagation, hence the influence of the decoupled network architecture is less related. In other words, the performance among models may not be optimal but is still comparable as all models share same settings and the differences are brought by propagation. 
Previous works of pre-propagation models already provide abundant exploration on different network depth and width \cite{Bojchevski2020,Chen2020a}. Hence we mainly adapt the settings in their major experiments to produce comparable results.

\fi

\subsection{Effect of \rename{}}
\label{ssec:featpush}
\ifx\istech\undefined
    \blue{To examine the \rename{} technique utilized in our model, we conduct ablation study to compare the precomputation performance of \tname{} by applying the \rename{} as in \refalg{alg:featreuse} or without reusing features and only full-precision \pprname{} computation. 
    We choose the Reddit dataset to generate artificial feature matrices with different dimensions $F$ to evaluate the feature-oriented optimization. The results of average times and testing accuracies over these feature matrices are given in \reftab{tab:reuse}. }
    
    \blue{By comparing the relative speed-up in \reftab{tab:reuse}, we state that \rename{} substantially reduces the precomputation time for different node feature sizes. When the number of features increases, the algorithm benefits more acceleration from adopting the feature optimization scheme, and achieves up to $1.6\times$ speed-up compared to solely applying \pprname{} propagation. }
    \blue{Meanwhile, \rename{} causes no significant difference on effectiveness as minor accuracy fluctuations are under the error bound of repetitive experiments. }
    Notably, the similar experiment on Papers100M shows that applying \rename{} saves more than 10 seconds solely in precomputation, which is a solid improvement for graphs with large node sizes. More detailed results can be found in \cite{techreport}.
    \input{floats/tab_reuse}
\else
    \blue{To examine the \rename{} technique utilized in our model, we conduct ablation study to compare the precomputation performance of \tname{} by applying the \rename{} as in \refalg{alg:featreuse} or without reusing features and only full-precision \pprname{} computation. 
    We choose the Reddit dataset to generate artificial feature matrices with different dimensions $F$ to evaluate the feature-oriented optimization. The results of average times and testing accuracies over these feature matrices are given in \reftab{tabr:reuse}. }
    
    \blue{By comparing the relative speed-up in \reftab{tabr:reuse}, we state that \rename{} substantially reduces the precomputation time for different node feature sizes. When the number of features increases, the algorithm benefits more acceleration from adopting the feature optimization scheme, and achieves up to $1.6\times$ speed-up compared to solely applying \pprname{} propagation. }
    \blue{Meanwhile, \rename{} causes no significant difference on output embedding values and effectiveness, as minor accuracy fluctuations are under the error bound of repetitive experiments. }
    \vspace{1.5ex}
\begin{table}[hbp]
  \caption{Effect of \tname{} with and without \rename{} on precomputation time ($\si{s}$), testing accuracy (\%), and average embedding value difference ($\times 10^{-7}$) for Reddit dataset with different feature dimensions $F$. }
  \label{tabr:reuse}
  \centering
  \small
  \begin{tabular}{@{}cc|rrrr@{}} \toprule
    \textbf{ } & \textbf{Feature} & 
        $\bm{F=100}$ & $\bm{F=200}$ & $\bm{F=400}$ & $\bm{F=602}$ \\
    \midrule
    \multirow{3}{*}{\textbf{Pre. Time}} & 
       w/o \textsc{Reuse} &
         0.46 &  0.93 &  1.83 &  2.84 \\
     & w/ \textsc{Reuse} &
         0.35 &  0.67 &  1.30 &  1.85 \\
     & Speed-up &
        133\% & 138\% & 141\% & 155\% \\
    \midrule
    \multirow{3}{*}{\textbf{Accuracy}} & 
       w/o \textsc{Reuse} &
         27.7  &  31.9  &  37.0  & 40.5 \\
     & w/ \textsc{Reuse} &
         27.8  &  31.7  &  36.7  & 40.3 \\
     & $\Delta$ &
        $+0.1$ & $-0.2$ & $-0.3$ & $-0.2$ \\
    \midrule
    \multicolumn{2}{c|}{\textbf{Embedding Difference}} &
        2.10 & 1.50 & 1.65 & 1.56 \\
    \bottomrule
  \end{tabular}
\end{table}

\fi

\vspace{-0.5ex}
\section{Conclusion}
\label{sec:conclusion}
In this paper, we propose \tname{}, a scalable Graph Neural Network algorithm with feature-oriented optimizations. Our theoretical contribution includes showing the \tname{} model has a sub-linear complexity that efficiently scale-up the graph propagation by \pprname{} and \rename{} algorithms. We conduct extensive experiments on various datasets to demonstrate the scalability of \tname{} in precomputation, training, and inference. Our model is efficient to process billion-scale graph data and achieve up to $100\times$ faster than the current state-of-the-art scalable GNNs in precomputation, while maintaining comparable or better accuracy. 

\bibliographystyle{ACM-Reference-Format}
\bibliography{main}

\end{document}